# *Integral Biomathics*:

# A Post-Newtonian View into the Logos of Bios

**Author:**           Plamen L. Simeonov
**Contact address:**  Wilhelmstrasse 91
                     D-10117 Berlin, Germany
**Telephone:**        +493025329555
**Fax/UMS:**          +4930484988264
**Email:**            plamen.l.simeonov@gmail.com

*Integral Biomathics*: A Post-Newtonian View into the Logos of Bios

**Plamen L. Simeonov**

plamen@simeio.org

**Abstract**

This work is an attempt for a state-of-the-art survey of natural and life sciences with the goal to define the scope and address the central questions of an original research program. It is focused on the phenomena of emergence, adaptive dynamics and evolution of self-assembling, self-organizing, self-maintaining and self-replicating biosynthetic systems viewed from a newly-arranged perspective and understanding of computation and communication in the living nature. The author regards this research as an integral part of the emerging discipline of nature-inspired or natural computation, i.e. computation inspired by or occurring in nature. Within this context, he is interested in studies which represent a significant departure from traditional theories about complex systems and self-organization, emergent phenomena and artificial biology. In particular, these include non-conventional approaches exploring the aggregation, composition, growth and development of physical forms and structures, autopoiesis along with the associated abstract information structures and processes.

This paper provides a critical review of the major assumptions which guide the development of modern computer science and engineering towards emulating biological systems. For this purpose, the author explores the potential and the virtues of biology to reshape contemporary science.

The goal of this survey is to discuss the present state of natural and engineering sciences in the light of a necessary paradigm change in the structure and methodology of research and deliver some insights for developing a new kind of integral science based on the principles for dynamic interdependence of the constituting disciplines and on the evolving relationships among them.

Word count: 248 (Limit: 250)

*Integral Biomathics*: A Post-Newtonian View into theLogos of Bios

**Plamen L. Simeonov**
plamen@simeio.org

> "I'm not happy with all the analyses that go with just classical theory,
> because Nature isn't classical… How can we simulate the quantum mechanics? ...
> Can you do it with a new kind of computer – a quantum computer?
> It is not a Turing machine, but a machine of a different kind."
>
> Richard Feynman, Simulating physics with computers, 1981.

**1. Introduction**

This work is an attempt at a state-of-the-art survey of natural and life sciences with the goal of defining the scope and addressing the central questions of an original research program. It is focused on the phenomena of emergence, adaptive dynamics and evolution of self-assembling, self-organizing, self-maintaining and self-replicating biosynthetic systems viewed from a newly-arranged perspective and understanding of computation and communication in living nature. The author regards this research as an integral part of the emerging discipline of *nature-inspired* or *natural computation,* i.e. computation inspired by or occurring in nature (Ballard, 1997; Shadboldt, 2004; MacLennan, 2005; Zomaya, 2006). Within this context, he is interested in studies which represent a significant departure from traditional theories about complex systems and self-organization, emergent phenomena and artificial biology. In particular, these include non-conventional approaches exploring the aggregation, composition, growth and development of physical forms and structures, autopoiesis (Maturana & Varela, 1980), along with the associated abstract information structures and processes.

This paper is not about a solution to a mathematical problem or an engineering implementation. Instead it explores the potential and the virtues of biology to reshape contemporary science in line with Popper's critical rationalism approach (Popper, 1934), yet in an extended autopoietic (Salthe, 1993; Kawamoto, 2000; Boden, 2000; Nomura, 2002 ff.) and non-Darwinist way. The author does not discuss the implications of Darwin's indubitable contribution to biology, because a detailed review of this subject goes beyond the scope and the focus of this paper. [Incidentally, the Darwinians have incorporated co-operation and symbiosis into the original theory, because sometimes they outcompete competition (Augros & Stanciu, 1995; Goldenfeld & Woese, 2007).] This survey also does not pay enough attention to development and evolution (Salthe, 1993). The author considers a return to these issues in future work. The purpose of this paper is not to present another constructivist approach in biological computation, but to survey and discuss the present state of natural and engineering sciences in the light of a necessary paradigm change in the structure and methodology of research (Kuhn, 1962; Kuhn, 2000). Hence, this work is more about *discovering* and *understanding* living systems rather than about anticipating an a priori design. Its primary goal is to ask the right questions in the context of an ecosystem of interacting biological entities and human-made artefacts. It is the author's intention to provide a critical review of the major assumptions which guide the development of modern computer science and engineering towards emulating biological systems. The goal of this survey is to deliver some insights for developing a new kind of integral science based on the principles of dynamic interdependence of the constituent disciplines and on the evolving relationships among them.

This study was motivated by previous research in system design and network engineering (Simeonov, 1998; Simeonov, 1999a/b/c; Simeonov, 2002a/b/c), where the limits of contemporary information technology and multimedia communication systems were identified and a novel approach towards autopoietic networking (Wandering Logic Intelligence, WLI) was proposed. Recent developments in information and communication technology towards ultimate dispersion and virtualization of resources support this approach. Nevertheless, there is a lot of work to be done on the road to a truly integrative science, where mindset habits and dogmas will give way to unbounded creativity and innovation.

The present work continues the above line of research towards deeper understanding of biological phenomena such as emergence and organisation in a holistic manner as seen in *relational biology* (Rashevsky, 1954 ff.; Rosen, 1958a/b ff.; Martinez, 1964) and *biological relativity* with multilevel biosystems (Rashevsky, 1954) which show no privileged level of causation (Nottale, 1993; Auffray & Nottale, 2008; Nottale & Auffray, 2008; Noble, 2008). The author's ultimate objective is to unify classical mathematical biology with biomathics on the way to genuine biological system engineering. *Biomathics* or biological mathematics is defined as the study of mathematics as it occurs in biological systems. In contrast, mathematical biology is concerned with the use of mathematics to describe or model biological systems, (Rashevsky, 1940). Therefore, this study is going to be carried out both from the perspective of traditional (analytic) life and physical sciences, as well as from that of engineering (synthetic) sciences. In this regard, the author's attempt differs from most present day efforts of biomimetics in automata and computation design such as neuromorphic engineering (Mead, 1990; Jung et al., 2001) to develop autonomic systems by emulating a limited set of "organic" features using traditional mathematical methods and computational models.

This approach is about asking *what is computing and cognition*, and about understanding where the biological imperatives for them come from and lead to, rather than being about replicating some aspects of them. Therefore, the author distances himself from the constructivist state-based approaches to human and machine intelligence (Turing, 1948, 1950a, 1951a/b; Braitenberg, 1986; Brooks, 1999; Porr & Wörgötter, 2003a/b). He also declines recently elevated appeals in favour of "biologically uninspired computer science" (Teuscher, 2006) for the plain reason that scientific disciplines cannot develop isolated from each other within the context of evolutionary epistemology of theories (Kuhn, 1962-1963; Toulmin, 1967-1972; Campbell, 1974; Lorenz, 1977; Popper, 1972-1984; Hull, 1990). Furthermore, it is essential to note that classical information theory (Shannon, 1948) should be developed along the same line of research in order to obtain an authentic picture of natural biological systems (Rashevsky, 1955a) that will enable the creation of artificial ones. This viewpoint has certainly become an important issue in the design of complex networked systems deploying large numbers of distributed components with dynamic exchange of information in the presence of noise and under power and bandwidth constraints in the areas of telecommunications, transport control and industrial automation. Therefore, the author regards it as a part of his approach.

To address these critical issues, researchers pursue the improvement and unification of classical theories such as those of control, thermodynamics and information. For instance, Allwein and colleagues propose the integration of Shannon's general quantitative theory of communication flow with the Barwise-Seligman general qualitative theory of information flow to obtain a more powerful theoretical framework for qualitative and quantitative analysis of distributed information systems, (Allwein et al., 2004). Other authors are concerned with important theoretical issues such as the estimation of reliable noisy digital channel state (Matveev & Savkin, 2004) and the treatment of data density equilibrium analogous to thermodynamic equilibrium, (Kafri, 2006, 2007). Some works also introduce the physics of information in the context of biology and genomics, (Adami, 2004). However, what is important for the design of naturalistic systems is the perception of signalling and information content (including their processing and distribution) from the perspective of biological systems (Miller, 1978) and in correlation with autonomous regulation of power consumption and other life maintaining mechanisms. This topic has not been addressed sufficiently by present research in either natural or artificial systems. Therefore, it should become an integral part of the models and methods for approaching naturalistic computation.

This survey is organized as follows. The work develops along two planes or conceptions of discourse, the *physical* or the 'realization' plane and the *logical* or 'abstract' one. [The author refers here to the first two spheres of Penrose's "Three Worlds Model" (Penrose 1995, p. 414; Penrose, 2004, p. 18), the physical world of phenomena and the Platonic mathematical world.] The following two sections review previous research in naturalistic computation along these planes. Section four is devoted to non-classical computation models beyond the Turing machine model and to the relations of bioinformatics and biocomputation within a new context.

Next, section five introduces the kernel part of this article, the integral approach to biological computation. Section six discusses its impact on system modelling, while section seven deals with the overall implications in natural and life sciences. Finally, sections eight, nine, and ten provide summary, conclusions and outlook for research in Integral Biomathics. An epilogue in section eleven closes the survey. A reference list is provided in section twelve.

**2. The Physical Plane**

The *physical plane* lies within the domains of autonomous cellular automata (CA) and evolving complex systems such as autopoietic, autocatalytic and non-linear eco-networks. This area comprises classical systems and 'discrete' automata theories endorsed by system-theoretical (Rössler, 1971), artificial intelligence (AI) and artificial life (ALife) approaches such as evolutionary computation (Fogel et al., 1966; Rothlauf et al., 2006), machine learning (Mitchel, 1997), synthetic neural networks (Dyer, 1995; Smith, 2006) or adaptive autonomous agents (Maes, 1995) for transducing knowledge from biology and related life science disciplines into computer science and engineering. Such systems and automata are often referred to as organic or bio-inspired (Păun, 2004).

The conceptual and theoretical foundations of these fields have been elaborated in earlier works on autonomous systems such as the physico-mathematical models of neural excitation and inhibition (Rashevsky, 1933, 1939) and the Boolean models of neural networks (Pitts, 1943; McCulloch & Pitts, 1943), as well as Turing's reaction-diffusion systems (1952), Moore's artificial living plants (1956a), sequential machines (1956b) and other machine models (1962), von Neumann's kinematic model and Universal Constructor (1966), Rosen's ($M$, $R$)-systems (Rosen, 1964, 1968a, 1971, 1972; Loute, 2006) and two-factor systems (Rosen, 1967, 1968b), Conway's game of 'Life' (Gardner, 1970-71), Arbib's self-reproducing universal automaton (1966) and self-replicating spacecraft (1974), Dyson's self-replicating systems (1970, 1979) and Drexler's bio-nanomolecular engines (1986, 1992). An extensive body of basic research in AI (Putnam, 1967; Searle, 1980; Fodor, 1980; Jackson, 1985; Minsky, 1988; Haugeland, 1989; Boden, 1990; Feigenbaum & Feldman, 1995; Chalmers, 1997; Nilsson, 1998; Minsky, 2007; Carter, 2007; Jones, 2008; Floreano & Mattiussi, 2008; Fodor, 2008; Johnson, 2008; Munakata, 2008; Russel & Norvig, 2009) and ALife (Levy, 1993; Brooks & Maes, 1994; Boden, 1996; Langton, 1997; Adami, 1999; Langton et al., 2003; Rocha et al., 2006; Komosinski & Adamatzky, 2009) goes along with these developments. Some of the systems, such as the simplified cellular automatons were realized in practice, (Codd, 1968; Winograd, 1970; Morita & Imai, 1995).

However, discrete mathematics has its application limits in modelling and emulating biological systems. Richard Feynman came to a similar conclusion that classical computers are inappropriate for simulating quantum systems (1982, 1985). Apart from maintaining living functions, biological systems demonstrate complex computational mechanisms. The author holds that only artificial or hybrid biosynthetic systems exhibiting a behaviour which is characteristic to that of natural organic systems could truly reflect the essence of biological computing. Therefore, he argues that this area of research needs to be placed on broader foundations which more adequately reflect the emergence and organisation of artefacts and processes in nature than modern discrete automata and computation approaches. This work's intention is to expand the scope of research on the emergence and organisation of living forms (Thompson, 1917; Bertalanffy, 1928; Rashevsky, 1944 ff; Franck, 1949; Rosen, 1991; Miller, 1978; Thom, 1989; Miller & Miller, 1990; Luisi, 2006) and their networks of production within a broader discourse and beyond the classical state automata theory and traditional models in life and physical sciences.

There is a basic dichotomy between the real and artificial forms of life: *reductionism*. Living systems are open, self-organising systems that have the special characteristics of life and interact with their environments by means of information and material-energy exchange (Miller & Miller, 1990), whereas artificial life is closed and can be ultimately reduced to information processing (Levy, 1993).

Yet, there have been a number of artificial life techniques using computational models and algorithms adopted from life science disciplines such as genetics, immunology and neurology, as well as molecular and evolutionary biology (Ray, 1995; Schuster, 1995). Among the most prominent examples of artificial life systems are *Tierra* (Ray, 91) and *Avida* (Adami, 99). Other well-known references are the SCL model (McMullin, 1997a), self-assembling cells (Ono & Ikegami, 1999-2000), self-assembling DNA nanostructures (Reif et al., 2001; Reif et al., 2007) and self-assembling lipid bilayers (Rasmussen et al., 2001). Recently, artificial chemistry approaches have been receiving much attention (di Feroni & Dittrich, 2002; Hutton, 2002; Matsumaru et al., 2006). In particular, experimentation on design and implementation of living systems from scratch is becoming an intensive research area (Bedau, 2005; Bedau, 2006). Good intermediate reports of the field are given in (McMullin, 2004) and (Guruprasad & Sekar, 2006).

As yet, the disciplines of artificial life and chemistry still hide many open issues (Bedau et al., 2000; McMullin, 2000a) such as the controversial matter about the possibility of engineering molecular and self-replicating assemblers (chemistry with/without mechanics or the Smalley vs. Drexler debate) in nanotechnology (Reif et al., 2001; Baum, 2003; Freitas & Merkle, 2004; Reif & LaBean, 2007). One of them refers to a basic question and probably the most known effort to explain life in general, the *autopoietic theory* (Varela et al., 1974; Maturana & Varela, 1980) which has influenced various sciences including biology and sociology for the past 30 years (Kneer & Nassehi, 1993).

One key aspect of biology is to understand the non-mechanistic nature of the world. Our models of the physical world are crisp and fully explicit following reductionist patterns, while the world is in various degrees implicit and vague. Autopoiesis provides therefore a good approximate model for understanding the organization and evolution of natural living systems. The question in artificial life research is, however, whether this model can serve as a base for creating synthetic organisms that mimic real ones. So far autopoiesis has been used as the base for a number of computational models, simulations, engineering and architecture solutions (Zeleny & Pierre, 1975; Zeleny, 1978; McMullin & Varela 1997; Cardon & Lesage 1998; Ruiz-del-Solar & Köppen, 1999; McMullin & Groß 2001; Simeonov, 2002; Kawamoto, 2003; Beer, 2004; Keane, 2005). However, autopoietic theory has obviously failed to provide a consistent view on the spontaneous organization of living systems in a quite early stage of their development (Zeleny, 1980). Since then, it has been subject of various criticisms and controversial discussions. Thus, a key point in understanding the mechanisms of self-assembly and self-organisation in living systems is the notion of organisational *closure* (McMullin 2000b). According to Maturana and Varela, an autopoietic system is not only one that is (a) clearly separated from its environment by a boundary, but also one that has (b) an internal organisation capable of dynamically sustaining itself (including its boundary). It is yet not clear how stringently this definition should be taken. Today, autopoiesis appears to be still the kind of "theory-at-work" which is very general and undifferentiated both in terms of mathematical formalization and technical implementation. The following two sections illustrate this.

## 2.1 The Formalization Gap

In his book "Life Itself" (Rosen, 1991) Robert Rosen presents a category theoretical framework for formalization of living systems he studied over three decades, (Rosen, 1958 ff.). There he poses what the author calls the *Fundamental Question of Artificial Life*. According to Rosen's conclusion, living systems, which are essentially metabolism/repair (*M*, *R*)-systems (Rosen, 1964, 1968a, 1971, 1972), are not realizable in computational universes. If Rosen were right, his conclusion could mean that Artificial Life cannot exist at all or at least in computational spaces as they are known now. Since autopoiesis also implies that strong ALife is impossible (Boden, 2000), it could be the case that the entire ALife research is going in the wrong direction today. Letelier et al. (2004) analysed (*M*, *R*)-systems from the viewpoint of autopoietic theory. They provided an algebraic example on defining metabolic closure while suggesting a relationship to autopoiesis.

In a series of works (Nomura, 1997, 2002, 2006), reflecting the contributions of Rosen (1972, 1991), Luhman (Kneer & Naseli, 1993) and Kawamoto (1995, 2000), Nomura reviewed the formal roots of autopoiesis in the light of category theory (Mitchell, 1965) and proposed a mathematical model of autopoiesis based on Rosen's definition (1997). After having examined some of the central postulates of the autopoietic theory, Nomura comes to the conclusion that *previous research has not delivered an unambiguous description of the phenomenon* (♣ [1]) and proposes a more general and strict formal definition.

Another category theoretical argument and revision of Rosen's theorem was provided by Chu and Ho (Chu & Ho, 2006). These authors review the essence of Rosen's ideas leading up to his rejection of the possibility of simulating real artificial life in computing systems. They argue that the class of machines Rosen distinguished from closed systems is not equivalent to a realistic formalization of ALife and conclude that Rosen's central proof, stating that living systems are not mechanisms, is wrong. Yet, Rosen himself warned in his book: "there is no property of an organism that cannot at the same time be manifested by inanimate systems" (*Life Itself*, page 19). As a result, Rosen's claim remains an open issue, or perhaps not.

The conclusion that some of Rosen's central notions were probably ill defined provides some interesting theoretical concerns which deserve further investigation. Having him apparently failing in his own proof does not change the matter at all. From the viewpoint of contemporary logic, the author cannot take for granted that organisms are mechanisms and that they can be constructed in ways used to build machines. Indeed, "Rosen's model did NOT fail regarding particular aspects of some natural system which can have mechanical representations. It did fail to represent whole *developing* systems" (Salthe, 2007, pers. corresp.). The last issue is central for a comprehensive model of living systems.

In a recent paper (2007), Nomura analyses the possibilities of algebraic description of living systems and clarifies the differences between the aspects of closedness required in (*M, R*)-systems and autopoiesis. He discovers two essential differences between autopoietic and (*M, R*)-systems. The first one is the difference in their forms of closedness under entailment of the components and categories required for the description of closedness. The second one is the distinction between organization and structure. Nomura points out that the first difference depends on the assumption that completely closed systems, modelled as an isomorphism from the space of operands to the space of operators, are necessary conditions of autopoiesis. However, this requirement has not yet been proved in a mathematically strict way. Furthermore, the definition of autopoiesis itself deserves a special attention. There were differences in the interpretation of autopoiesis between Maturana and Varela. When Varela collaborated with McMullin on computational models of autopoiesis, the original algorithm was revised within the same year (McMullin, 1997b; McMullin & Varela, 1997). The author assumes that Varela may not have considered the implications of autopoietic theory on the axiomatisation of discrete mathematics for modelling biological processes.

Summarizing all these facts, the following two conclusions can be made: (i) a more precise and formal definition of autopoiesis and its relationship to (*M, R*)-systems, as outlined by (Nomura, 2007), is required; (ii) novel mathematical techniques and tools which adequately describe and simulate biological processes, (Rashevsky, 1954, 1960a, 1961) are required. In other words, in order to make progress in this area, it is necessary to either provide further means for formalization, or invent new formal approaches that best suit the original definition, or to redefine (refine, extrapolate) the theory itself, such as the tangible approach provided in (Beer, 2004). Rosen tried to distinguish life systems from machines with his original definition. Although his proof was incomplete, Chu and Ho identified the importance of Rosen's idea. Nomura also agrees with them in this point (2007). Therefore, in order to provide the engineering base for artificial organisms and systems in the physical space, it is advisable to further investigate the necessary conditions for modelling characteristics of living systems and provide more stringent definitions of life systems and machines based on Rosen's attempt.

---

[1] With the symbol ♣ the author denotes the anchor arguments of our conclusion about Integral Biomathics henceforth.

## 2.2 The Implementation Gap

In the C5 database product presentation, Stalbaum makes the point that it is necessary to differentiate between the implementation of computational autopoiesis as proof of concept ('computational autopoiesis is possible') and the potential practical applications of such a system (Stalbaum, 2007). He reckons that Varela's early work makes a strong case for the former (McMullin & Varela, 1997), but that designing autopoietic automata implementing computing applications such as database (self-) management systems is a substantially different problem.

Stalbaum states that the challenge in finding or engineering congruency between autopoietic systems and problems that yield solutions is *enormous*. Indeed, the demonstration that autopoiesis can be used for computational and communication processes using a minimal implementation, e.g. of an artificial chemistry model, is relatively simple. However, the truthfulness of computational autopoiesis does not necessarily imply that autopoiesis can be effectively implemented to perform work. This is because the internal purpose of an autopoietic system is restricted to the ongoing maintenance of its own organization. Yet, this goal becomes a problem for anyone intending to use autopoiesis for the purpose of computation as it is known, i.e. to deliver an output result from a given number of inputs within a limited number of *steps*.

In fact, computation might be a by-product of ongoing structural coupling (a posteriori) between a collection of autopoietic elements such as neurons and their environment, but it cannot be deterministically defined as a purposeful task for the solution of a specific problem or class of problems in the way expected from today's computational and engineered systems.

Decision making is not *the fundamental intent* of biological computing. In other words, the natural drift inherent in living systems cannot be counted on to directly solve problems that do not primarily serve for *conservation, adaptation* and *maintenance* of those systems. In addition, the multiple orders of structural coupling in autopoiesis define an even more complex picture of the interacting units. In this respect, autopoietic computing is *analogous* to both associative computing (Wichert, 2000) and quantum computing (Feynmann, 1982-1985). [The term "analogous" is used with multiple meanings in this context: i) "similar" as in modern English, and ii) "relational" (to emphasise the approaches of Thompson, Rashevsky and Rosen), i.e. *proportional* (Latin: *rational*), or following the principle of mediation in the Pythagorean sense according the Greek lexis (Guthrie, 1987). In anatomy, organic structures are considered to be *analogous* when they serve similar functions, but are not evolutionarily related. Plato and Aristotle used even a wider notion of "analogy" meaning not only a "shared relation", but a "shared abstraction" such as an idea, a pattern, an attribute or a function (Sheley, 2003). Analogous is also a synonym for continuous, the *antonym* of discrete, computing. It demonstrates an over layered potential multiplicity of results which becomes apparent at the very moment of system interrogation.]

The following example illustrates more vividly the problem of implementing artificial biology using conventional computing techniques. A bottom-up synthesis approach, Substrate Catalyst Link (SCL), built on the concept of evolving autopoietic artificial agents (McMullin & Gross, 2001) was integrated within a top-down analytical system, Cell Assembly Kit (CellAK), for developing models and simulations of cells and other biological structures (Webb & White, 2004). The original top-down design of the CellAK system was based on the object-oriented (OO) paradigm with the Unified Modelling Language (UML) and the Real-Time Object-Oriented Methodology (ROOM) formalisms, with models consisting of a hierarchy of containers (ex: cytosol), active objects with behaviour (ex: enzymes, lipid bilayers, transport proteins), and passive small molecules (ex: glucose, pyruvate). Thus, the enhanced CellAK architecture comprised a network of active objects (polymers), the behaviour of which *causally* depended partly on their own fine-grained structure (monomers), where this structure was constantly changing through interaction with other active objects. In this way, active objects influence other active objects by having an effect on their constituent monomers.

The enhanced tool was validated quantitatively vs. GEPASI (Mendes, 1993) and demonstrated its capability for simulating bottom-up synthesis using the cell bilayer active object. Web and White claim that this result clearly confirms the value of agent-based modelling techniques reported in (Kahn et al., 2003). However, there is a major difficulty in implementing this method for more complex organic structures than lipid bilayers such as enzymes and proteins (Rosen, 1978). This is because amino acids that compose proteins are coded in the DNA; their order results in a folding 3D shape which is of crucial importance. Therefore, the behaviour of a protein is an extremely complex function of its fine-grained structure. This turns the design and validation of artificial biological structures, such as pharmaceuticals, using conventional computing techniques into a problem of polynomial complexity.

The author does not believe that significant progress can be made in this area without a paradigm change. Taking into account the above arguments about formalization and implementation, he exposes the need for a new broader and unifying automata theory for studying 'natural' and artificial living systems, as well as the combination of both, the cybernetic organism (*cyborg*) or the hybrid between an animal/plant and a machine. Furthermore, the author demands a new kind of unifying and *real* Artificial Intelligence, the expected 'quantum leap for AI' (Hirsch, 1999), that goes beyond the pioneer days heuristics and hypothesis-driven modelling by placing itself much closer to the essence of living systems, and closer to the nature of the underlying processes in both organic and inorganic system complexes. The logic implied here is that modern science demands computation techniques and architectures which are biological processes and systems themselves.

## 3. The Logical Plane

The second plane for research the author is interested in is the abstract or *logical plane* in the Pythagorean sense of the word (Guthrie, 1987), i.e. one that implies analogy or geometrical proportion (Fechner, 1849; Thompson, 1917, p. 269), but not perceivable as plain composition (Ehrenfels, 1890), a relation that is generalized as a law, habit, principle or basic *pattern* of system organization (Gierer & Meinhardt, 1984; Wolfram, 2002; Barnsley, 2006). The essential distinction of this conception from Hilbert's pure syntactic definition is the inclusion of semantics or context-related information which allows the multiplicity of interpretations, a characteristic typical for biological, social and eco-systems. The logical plane is concerned with the development of new *integral computational paradigms* that reflect the emergence and organization of information for living systems in a more adequate way than traditional formal approaches based on binary logic and the Church-Turing thesis, (Church, 1932-41; Turing, 1936-51). This plane represents an idealized, but not necessarily unique, model for computation, where representations are formal, finite and definite (MacLennan, 2004).

Exposing these characteristics itself, the Church-Turing theory of *discrete states* excludes by definition alternative approaches to computation. Even worse, as it has been successful for the past 70 years, the Church-Turing model evokes the conviction in the majority of contemporaries that it is universal, as demonstrated in (Shannon, 1956), and that there is no other option for computation at all. Thus, binary logic remains a good working, and of course, economic model for everyday discourse, just as Newtonian celestial mechanics was useful for global navigation until the arrival of Einstein's theory of relativity (1905-1915) which made interstellar journeys conceivable. Yet, the quest for alternative approaches to computation continues.

Therefore, the author exposes the need for new theories of computation in order to understand tough issues in science such as the emergence and evolution of brain and thought.

In his books "The emperor's new mind" (Penrose, 1989) and "Shadows of the mind" (Penrose, 1994), Roger Penrose's main theme was the understanding of mind from the perspective of contemporary physics and biology. While discussing the non-computational physics of mind, he posed indirectly what the author calls the *Fundamental Question of Strong Artificial Intelligence* ["...according to strong AI, the computer is not merely a tool in the study of the mind; rather, the appropriately programmed computer really is a mind", (Searle, 1980)]: Penrose claimed to prove that Gödel's incompleteness theorem, stating that Number Theory is more complex than any of its formalizations (Gödel, 1931-34), implies that human thought cannot be mechanized. In fact, Penrose does not actually use Gödel's theorem, but rather an easier result inspired by Gödel, namely, Turing's theorem that the halting problem is unsolvable.

Penrose's key idea was essentially the same as that of the philosopher J. R. Lucas (Lucas, 1961) who argued as follows. Gödel's incompleteness theorem shows that, given any sufficiently powerful formal system, there is a true sentence which the formal system cannot prove to be true. But since the truth of this unprovable sentence is proved as part of the incompleteness theorem, humans *can* prove the sentence in question. Hence, human abilities cannot be captured by formal systems made by humans. This corresponds to the eye's *blind spot paradox* which in the author's view should be regarded rather as a principle in biological systems. The blind spot in an eye's vision field is caused by the lack of light-detecting photoreceptor cells on the optic disc of the retina where the optic nerve passes through it. Usually, this 'defect' is not perceived, since the brain compensates the missing details of the image with information from the other eye.

The question here is again, as in Rosen's thesis about Artificial Life (Rosen, 1991), see Section 2.1, whether this proof is correct or not. If Penrose and Lucas are right, their conclusions might lead to the interesting result that Strong Artificial Intelligence cannot be realized at all or at least not in computational spaces as they are currently known.

Lucas's standpoint was already criticized by Benacerraf (1967). Some of Penrose's technical errors were pointed out in (Boolos et al., 1990) and (Putnam, 1995). LaForte and colleagues review Penrose's arguments and demonstrate, following Benacerraf's line of thought that they depend crucially on *ambiguities between precise and imprecise definitions of key terms* (♣♣), (LaForte et al., 1998). The authors show that these ambiguities cause the Gödel/Turing diagonalisation argument to lead from apparently *intuitive* claims about human abilities to paradoxical or highly idiosyncratic propositions, and conclude that any similar argument will also fail in the same ways. This is a similar situation to the contra-proof of Rosen's thesis discussed above, (Nomura, 2006; Chu et al., 2006).

In fact, the arguments of Lucas and Penrose have the same foundation as Rosen's discussion about the entailment of formal systems (Rosen, 1991, pp. 47-49), stating that "from the standpoint of the formalism [assuming here Gödel's thesis], *anything,* that happens outside is accordingly *unentailed*". However, Aristotle's fourth category about the Final Cause (a proposition that requires *something* of its own effect) which appears to violate the unidirectional (temporal) flow from axioms to theorems, places (some) human abilities inside this new kind of *enhanced* formal system of life. Rosen finds a solution to this contradiction by postulating that final causation requires modes of entailment that are not generally present in formalisms. Following Rashevsky's concept of relational biology (Rashevsky, 1960b), he also suggests the possibility to *separate finality from teleology* by retaining the former while, for example, discarding the latter. A similar approach could be taken here to investigate the Lucas/Penrose arguments more precisely. Another solution may be placed entirely or partly on quantum mechanical foundations (entanglement).

The final word about strong artificial intelligence has not been said yet. The main reason is that "…a concerted effort in formalizing interdisciplinary ALife and biology research to achieve the common goal of studying and recreating life-like and living systems is lacking", (Guruprasad & Sekar, 2006).

For instance, fuzziness as simulacra of vagueness represents an essential characteristic of living systems. No matter how far fuzzy models and algorithms (Zadeh, 1965, 1968) have gone, there is still need for 'logic of vagueness' (Peirce, 1869a; Brock, 1969; Brock, 1979). One specific problem is that computer scientists and engineers consider fuzziness solved within the Turing machine model. Yet, this kind of "unscharfe" representation is not quite the same as what Peirce had in mind (Rescher, 1979; Esposito, 1980). There is fuzziness, then second order fuzziness (where the set itself is fuzzy as well as the membership (Klir, 1991; Wang & Klir, 1992; Klir, 2000), and then, far down the road is vagueness. The author will return to this topic in future work.

Whatever result the polemics around computation and human intelligence may deliver, as in the case with artificial life, the author cannot take for granted that human thought is mechanistic, i.e. formal, finite and definite, and that it can be constructed in the way used to build computers so far. "The conditions sought by the mechanistic explanation of nature explain only part of the contents of external reality. This intelligible world of atoms, ether, vibrations, is only a calculated and highly artificial abstraction from what is given in experience and lived experience." (Dilthey 1991, p. 203)

For instance, human *intuition* is an example of a creative act of obtaining knowledge from an "unidentifiable" source which does not require a formal proof. It is a process which cannot be expressed by formal means. Despite intuition not being well understood, it plays an essential role within human reasoning, a fact well known to Hilbert, Gödel, Poincaré, Weyl and Bohr who were influenced by the works of Kant (Carson & Huber, 2006). Thus, often an engineering model or a scientific theory comes into being as an *analogy*, i.e. by means of involving semantics, of another system description. For instance, it can simply pop up *intuitively* during a talk or discussion with the persuasion that this is the optimal answer for the domain in question prior to logical reasoning (Whitehead, 1933).

In fact, the modern conception of reason, being associated with (formal) logic and sharply contrasted with intuition, is a "thin" one (Fricker, 1995), whereas intuition is considered to be the catalyst for theoretical changes in science (Kuhn, 1962). There are numerous examples in the scientific literature that confirm such common experiences and the Gödelian implicit truth-without-proof, so that the process of its discovery, or disentanglement of semantics into the axiomatic frame of a specific formalism, can be taken as objective reality and *principle* in living systems (Penrose, 1994, 2004).

Hence, there is no need of hypothetical theories about time travel and precognition; this is a different kind of awareness and computation from those that contemporary technology delivers today. Ultimately, from the standpoint of Aristotelian epistemology, it is not a question of what and how was something proved, but rather *why* did this happen. A reasonable answer may contain the assumption that the solution of a problem is delivered by the human system itself, as a *useful* response to an external stimulus.

Therefore, the author claims that computation which occurs in nature always involves semantics and cannot be expressed within formalisms in purely syntactic terms. Although this argument is relatively weak in physical systems, an adequate picture of biological phenomena requires semantics. In the author's view, new opportunistic theories of computation should basically regard computation as major property of living matter and be able to develop their own principles within their specific domains of application. In this respect, the presented approach to the logical plane conforms with MacLennan's one in his definition of computation as a physical process for the abstract manipulation of abstract objects, the nature of which could be discrete, continuous or hybrid (MacLennan, 2004, 2006). Besides, the introduction of broader and integral concepts for computing is supported by such arguments as the fact that a computer's functioning is based on the state superposition principle and can be realised both with classical and quantum elements (Oraevsky, 2000).

Basically, MacLennan's investigation into these alternative models of information processing led to a similar conclusion for computation as those of Feynman for quantum systems (1982, 1985) and Rosen for artificial biology (1991, 1999). MacLennan argues that conventional digital computers are inadequate for realizing the full potential of natural computation, and therefore that alternative and more brain-like technologies should be developed. In his view, shared by the author, the Turing Machine model is unsuited to address the class of questions about natural computing (MacLennan 2003a). Furthermore, MacLennan argues that *natural* computation which occurs in natural neural networks (Freeman & Skapura, 1991; Haykin, 1998) requires continuous and *non-Turing* models of computation (MacLennan 2003b), although these processes cannot be regarded as 'computation' in terms of the Turing machine definition (Turing, 1937).

Yet, Turing's thesis is not the ultimate verdict about computing. The reader need only to refer to Euclid's geometry in a historical context. There are numerous examples about how new generalizations in mathematics and physics emerged out of apparent axioms and postulates. By shifting the perspective, the latter turned out to be conceptual deadlocks as they were not able to deal with new ideas or factual observations.

When the new theories were finally reconciled with the established world order, the apparent 'paradoxes' turned out to be new, more general facts and the old frame of thought became a special case within the new one (Rosen, 1991). It is a well known fact that science requires sometimes a few iterations of denial and rediscovery over the centuries until a new idea or paradigm is accepted by the dominating majority (Whitehead, 1933; Popper, 1934; Kuhn, 1963; Sacks, 1995). [Max Planck: "A new scientific truth does not triumph by convincing its opponents and making them see the light, but rather because its opponents eventually die and a new generation grows up that is familiar with it." (Planck, 1949) This citation is known as *Planck's Principle* or Planck's Other Law and represents a central aspect of the evolutionary epistemology of theories (Toulmin, 1967; Popper, 1974).]

The situation with the dominant computing paradigm today, the Turing Machine (TM) model, is similar. This concept leads to a discrepancy when applying computer science methods in systems biology. Indeed there is nothing wrong with the TM model, as long as it is exploited within its *frame of relevance* which deals with questions about formal derivability and the limits of effective calculability (MacLennan, 2006). Yet, when computing moves to the domain of biology, it realizes that it does not have the mathematics to describe living systems (Nomura, 2007), p. 25.

A major paradox with adopting the TM model for natural computing was pointed out in (MacLennan, 2004). Accordingly, formal logic considers a function computable if for any input data the corresponding output would be produced after finitely many steps, i.e. a proof can be of *any* finite length. For the sake of completeness and consistency, the TM model imposes no bounds on the length of the individual steps and on the size of formulas, so long as they are finite.

The concept of time assumed in the TM model is not discrete time in the familiar sense that each time interval has the same duration. Therefore, it is more accurate to call TM time a *sequential time* (MacLennan, 2006). Since the individual steps are of indefinite duration, there is no use to *count* the number of steps to translate that count into real time. Temporal formalisms based on the TM model such as TLA (Lamport, 1994, 2002) do not change this situation. Consequently, the only reasonable way to compare the time required by computational processes is in terms of their asymptotic behaviour. Therefore, once the observers have chosen to ignore the *speed* of the individual steps, all they have as complexity measure is the *size* of the formulas produced during the computation or the growth degree of the *number of steps* with the size of the input.

In fact, Turingian sequential time is reasonable in a model of formal derivability or effective calculability, since the time *duration* required for individual operations was irrelevant to the research questions of formal mathematics.

However, this perspective leads to the result that any polynomial-time algorithm is 'tractable' and 'fast' (e.g. matrix multiplication ~ O ($N^3$)) and that exponential-time algorithms are "intractable"(e.g. Travelling Salesman), whereas problems which are polynomial-time reducible are virtually identical. This duration independent situation is indeed peculiar for natural systems. MacLennan ironically describes it as one where "an algorithm that takes $N^{100}$ years is fast, but one that takes $2^N$ nanoseconds is intractable." The reader cannot afford ignoring the duration of the steps in natural computing, where instant response has the value of survival for a living organism.

> "In nature, asymptotic complexity is generally irrelevant... Whether the algorithm is linear, quadratic, or exponential is not so important, as whether it can deliver *useful* results in required real-time bounds for the inputs that actually occur. The same applies to other computational resources. … it is not so important whether the number of neurons required varies linearly or with the square of the number of inputs to the net; what matters is the absolute number of neurons required for the numbers of inputs there actually are, and *how well* the system will perform with the number of inputs and neurons it actually has." (MacLennan, 2006)

In addition, adaptation and reaction to unpredicted stimuli, large scale cluster optimizations and continuity of input and output values in space and time, such as those occurring in neural networks, immune systems and insect swarms are other characteristics of the natural system that lie outside the scope of the TM model and cannot be addressed appropriately with traditional analytical approaches. Finally, evolvability and robustness, (Wagner, 2005), i.e. continuous development and effective (though not necessarily correct!) operation in the presence of noise, uncertainty, imprecision, error and damage complete the fragmentary set of characteristics of living systems, (Rashevsky, 1954, 1960a, 1965a; Miller, 1978; Miller & Miller, 1990).

Therefore, it is necessary to establish different criteria for evaluating biological computation based on its primary purposes. The following section discusses some alternative approaches to the TM model.

### 4. About Non-classical Computation Models

The term non-classical computation denotes computation beyond and outside the classical Turing machine model such as extra-Turing, non-Turing and post-Newtonian computation (Stannet, 1991). Representative approaches include Super-Turing and hypercomputation, as well as nano-, quantum, analog and field computation.

**Super-Turing computation** (Siegelmann, 1995, 1996a, 2003) is a synonym for any computation that cannot be carried out by a Turing Machine, as well as any (algorithmic) computation carried out by a Turing Machine. Super-Turing computers represent any computational devices capable of performing Super-Turing computation, e.g. non-Turing computable operations such as integrations on real-valued functions that provide exact rather than approximate results. In fact, Turing himself proposed a larger class of "non-Turing" computing architectures as alternatives including *oracle machines* (*o-machines*), *choice machines* (*c-machines*), and *unorganized machines* (*u-machines)*. Yet, he did not formalize them and they were discarded as unnecessary in the 1960s. It was assumed that (classical) Turing machine models could completely describe all forms of computation.

Though this narrow interpretation of the Church-Turing thesis contradicted Turing's assertion that Turing machines could only formalize *algorithmic* problem-solving, it became a dogmatic principle of the theory of computation (Wegner & Goldin, 2003). Thus, computer science adopted a computation model that paralleled the one of Newtonian physics and provided an acceptable, but weak theory of computation.

This fact that should have been realized earlier in natural sciences instead of researchers becoming used to computers as they were engineered so far (cf. Feynman's quote at the beginning of this paper). Turing did not anticipate that his sequential I/O model (called by him *automatic machine* or *a-machine*) would dominate problem solving in science over five decades (Eberbach et al., 2006).

In the 1970s and 1980s it became gradually clear that the increasing demands on computation power and communication between distributed systems require a theory of concurrency and interaction within a new conceptual framework, "not just a refinement of what we find natural for sequential [algorithmic] computing" (Milner, 1993). However, process algebra which shaped concurrent computation theory did not openly challenge TMs at that time. It simply placed interactions orthogonal to computation, rather than being a part of it.

Thus, models like CCS (Milner, 80), ACP (Bergstra & Klop, 1984) and CSP (Hoare, 1985) avoided direct confrontation with the establishment (Goldin & Wegner, 2005). Yet, they opened the door to novel approaches to computation beyond classical Turing machines. Among them are π-*calculus* (Milner, 1991 & 2004), *$-calculus* (Eberbach, 2000-2001), *Evolutionary Turing Machines,* – a more complete model for evolutionary computing than common Turing Machines, recursive algorithms or Markov chains (Eberbach & Wegner, 2003; Eberbach, 2005), – and *recurrent neural networks,* some of which use real numbers as weights (Siegelmann & Sontag, 1992; Siegelmann, 1993).

In the early 1990s, Brooks argued against the algorithmic approach of "old fashioned AI", thus favoring Super-Turing models based on interactions, learning and evolution (Brooks, 1991-1999).

Some developments of the TM in this way were:

- *Interaction Machines, IM* (Wegner, 1997-98): Turing machines cannot accept external input during computation, while interaction machines extend the TM model by input and output actions that support dynamic interaction with an external environment.

- *Persistent Turing Machines, PTM* (Goldin & Wegner, 1999; Goldin, 2000): multitape machines with a persistent worktape preserved between successive interactions; they represent minimal extensions of TM that express interactive behaviour characterized by input-output streams.

- *Interactive Turing Machines with Advice and Infinity of Operation (ITMAIO), Site and Internet Machines, SIM* (van Leeuwen & Wiedermann, 2000a/b).

The above models regard computation as an ongoing interactive process, possibly attended by distinct supporting activities such as the exchange of software and hardware (ITMAIO) rather than a purposeful functional transformation of input data into output data. Each one of them defines a specific solution frame. Thus, IM extend Turing machines with interaction to address the behavior of concurrent systems, promising to bridge these two fields (Goldin & Wegner, 2005), whereas PTM capture sequential interaction, which is a limited form of concurrency.

Yet, the computational power and expressiveness is not always clearly in favor of the Super-Turing interaction approaches. Sometimes they can be emulated by Turing machines. For instance, a site machine, which is a full-fledged computer equipped with a processor, a number of I/O ports and a potentially unbounded permanent read-write memory, can be modeled either by a Turing machine, or by a random access machine (RAM) equipped with an external permanent memory or by any other model of universal programmable computer. This holds also for an internet machine which is a finite, but time-varying set of site machines.

Nevertheless, although being criticized (Ekdal, 1999), interaction models are generally considered to allow desired extensions of the Church-Turing thesis into more general theses with expressiveness going beyond the one of Turing machines for the following reasons (Wegner, 1997):

- *formal evidence of irreducibility*: input streams of interaction machines cannot be expressed by finite tape, since finite representation can be dynamically extended by uncontrollable adversaries;
- *informal evidence of richer behavior*: TMs cannot handle the passage of time or interactive events during computation;
- *unpredictable and non-algorithmic*: interactive machines interact with an external environment they cannot control.

Exhaustive discussions of this issue are also given in (van Leeuwen & **W**iedermann, 2000a/b) as well as in (Wiedermann & van Leeuwen, 2002).

**Self-replicating cellular automata** (Neumann, 1966) are a special class of generic self-realizable computing architecture that the author assigns to the physical plane for the purpose of his dyadic model of an evolving intelligent "computational organism" discussed in the next section. In fact, cellular automata (CA) and some Super-Turing architectures belong to both planes since they can be an abstract concept and its physical realization. It should be kept in mind, however, that while "natural" organisms appear to be interaction machines (Neuman, 2008) they are certainly not self-replicating cellular automata. They cannot be "simply manufactured according to a set of instructions. There is no easy way to separate instructions from the process of carrying them out, to distinguish plan from execution" (Coen, 2000). Thus, internalism is omnipresent in the (living) world.

**Hypercomputation** (Copeland, 2004) studies models of computation that expand the concept of computation beyond the Church-Turing thesis and perform better than the Turing machine. It refers to various proposed methods for the computation of non-Turing computable functions such as the general halting problem. A good survey report in this area is given in (Ord, 2002), where ten different types of hyper-machines (infinite state TM, probabilistic TM, error prone TM, accelerated TM, infinite time TM, fair non-deterministic TM, coupled TM, TM with initial inscriptions, asynchronous networks of TM and O-machines, also classified as Super-Turing computation) are introduced and compared regarding their capabilities.

This review explains how such non-classical models fit into the classical theory of computation. Ord's central argument is that the Church-Turing thesis is commonly misunderstood. He claims that the *negative results of Gödel and Turing depend mainly on the nature of the theoretical descriptions of physical systems* (♣♣♣). This result is comparable with the ones in the discussions on the mechanization of *life* (♣) and *thought* (♣♣).

Other authors made similar conclusions (Rosen, 1962b; Kieu, 2002). The author also shares Ord's thesis that the Turing machine model is based on concepts conform to Newtonian physics which deals with closed conservative systems and on inadequate (from a biological viewpoint) mathematical abstractions such as absolute space and time, negligible power consumption and unlimited memory. Indeed, Newton's Principia have their precedents in the Aristotelian, Galilean and Cartesian schools.

On the other hand, quantum computation has already demonstrated (e.g. through such effects as quantum superposition, quantum entanglement or wave function collapse) that the feasibility of algorithms depends on the nature of the physical laws themselves. Thus, natural computation is relative just as biology is (Nottale, 1993; Auffray & Nottale &, 2008). Extrapolating *new* physical laws would automatically mean new computation approaches. This holds also in the case of biology, illustrated by such theories as Bohm's hidden variables interpretation of quantum theory (Bohm, 1952) and the Musèan *hypernumber* concept (Musès, 1972, 1976, 1994).

Therefore, to deal with the rising complexity of abstractions in biological computation, the author proposes the development of an evolving networked model for living systems (section 5, pp. 19-22) embracing the frame of *relational biology* set up by Rashevsky and Rosen (Rashevsky, 1954; Rosen, 1958a) on the base of Capra's *deep ecological theory* (Capra, 1997), – a synthesis of i) Bateson's Theory of Cognition (Bateson, 1972), ii) Prigogine's Theory of dissipative structures and life patterns (Prigogine, 1977), and iii) the Autopoietic Theory of biological networks and the processes defining and maintaining them (Maturana, Varela, 1980), – and addressing a new, inclusive theory of computation and automation based on the principles about emergence and self-organisation in general system theory (Bertalanffy, 1950 ff.), living systems theory (Miller, 1978) and systems biology (Kitano, 2002). The author does not see life as being "structured linguistically and organized communicatively" (Witzany, 2007, p. 9), but rather language, i.e. models, as structured bio-logically and communications organized *lively*, i.e. naturally in order to share not any but *vital* information.

The diversity of mathematical models for hypercomputers comprises some promising research fields in computer science, mathematics and physics including:
- a Turing machine that can *complete* infinitely many steps, (Shagrir & Pitowsky, 2003);
- an idealized analog computer (MacLennan, 1990; Siegelmann, 1996b), a *real* (numbers) computer that could perform hypercomputation if physics allowed in some way the computation with general *real* variables, i.e. not only computable real numbers;
- a *relativistic* digital computer working in a Malament-Hogarth space-time which can perform an infinite number of operations while remaining in the *past light cone* of a particular space-time, (Etesi &. Németi, 2002);
- a *quantum mechanical* system, but not an ordinary qubit quantum computer which uses an *infinite superposition of states* to compute non-computable functions, (Feynmann, 1982/85).

The above list illustrates how playful science approaches its research toolset, thus revealing an important epistemological aspect that somehow mimics mutation in nature. Currently all these devices are only theoretical concepts, but they may move some day to the physical (realization) plane and become everyday reality.

There is no fixed rule or reason in the quest for knowledge. Thus, empiricism goes hand in hand with rationalism. It should be kept in mind, however, that each new model in science represents a step in the joint evolution of experience and ideas. It is based on a set of assumptions (axioms) and an underlying former model representing a specific rationally-experienced interpretation or metaphor of the system at hand. For instance, introducing *real* numbers in an ideal analog computer in the second model above can be realized in two phases: i) by using a stochastic Turing machine that has binary coin as computation base and ii) which itself uses a *real* underlying process (but the machine never sees the process, only the binary coin). Such a "hybrid" analog machine can be Super-Turing too. So, the variables of real numbers do not have to be seen anywhere, but can be heavily hidden (Siegelmann, 1998).

On the other hand, a natural model of neural computing can give rise to hyper-computation as well. In (Siegelmann, 2003) an analog neural network allows for supra-Turing power while keeping track of computational constraints, and thus embeds a possible answer to the superiority of the biological intelligence within the framework of classical computer science. Siegelmann proposed it as standard in the field of analog computation, functioning in a role similar to that of the universal Turing machine in digital computation. An analog of the Church-Turing thesis of digital computation was also stated there, where the neural network takes place of the Turing machine.

**Nanocomputation** (MacLennan, 2006) involves computational processes with nano-devices and information structures which are not fixed, but in constant flux and temporary equilibria. It includes sub-atomic, molecular and cellular modes of computation such as *quantum* (Nielsen & Chuang, 2000; Bacon & Leung, 2007), *carbon nanotube* (Kong, 2007), *nanofluidic* (Marr & Munakata, 2007), *DNA* (Amos, 2005; Livstone et al., 2006) and *membrane computation* (Păun, 2003-2007).

A fundamental characteristic of nanocomputation is the microscopic reversibility in the device and information structures. This means that chemical reactions always have a non-zero probability of backwards flow. Therefore, molecular computation systems (Benenson & Shapiro, 2004) must be designed so that they (i) satisfy the requirements of an abstract molecular biology (Rashevsky, 1961c, 1962) and, (ii) accomplish their purposes in spite of such reversals. Furthermore, computation proceeds asynchronously in continuous-time parallelism and superposition. Also, operations cannot be assumed to proceed correctly and the probability of error is always non-negligible. Therefore, errors should be built into nanocomputational models from the very beginning. Due to thermal noise and quantum effects, errors, defects and instability are unavoidable and must be taken as given. Thus, new approaches which address these peculiarities are required for theorizing computational principles on a subatomic scale. Examples of nanocomputing devices include quantum logic gates and DNA chips.

**Analog computation** uses physical phenomena (mechanical, electrical, etc.) to model the problem being solved, by using uninterrupted varying values of one kind of physical parameter (e.g. water/air pressure, electrical voltage, magnetic field intensity, etc.) to obtain, measure and represent a goal function. A major characteristic is the operation on signals without conversion (sampling and integration) and in their natural continuous state.

Natural numbers and discrete computation have been used in everyday life since the abacus era. Yet, ABC, the first semi-electronic digital computer, pioneered by John Atanasoff in 1939 (Columbia University, 2005) announced the era of industrialization in computation and pushed forward the digital mode of calculation to dominate the market for the past 30 years due to technological advantage. Analog computers in turn have been also used since ancient times in agriculture, construction and navigation (Bromley, 1990). One of them is the Antikythera mechanism, the earliest known mechanical analog computer (dated 150-100 BC), designed to calculate astronomical positions, (François, 2006). When in 1941 Shannon proposed the first General Purpose Analog Computer (Shannon, 1941) as a mathematical model of an analog device, the Differential Analyser (Bush, 1931), this invention announced the age of electronic analog computers (Briant et al., 1960). From the very beginning, analog computers stepped in competition with their digital brothers, and initially outperformed them by optimally deploying electronic components (capacitors, inductors, potentiometers, operational amplifiers).

Analog computers have three major advantages over digital ones: i) instantaneous response, ii) inherent parallelism, and iii) time-continuity (no numerical instabilities or time steps). Until 1975 analog computers were considered to be unbeatable in solving scientific and engineering problems defined by systems of ordinary differential equations. However, their major disadvantage, the limited precision of results (3 to 5 digits), their size and price made them unsuitable for future applications with the advent of the transistor and the growing performance of integrated circuits in digital computers. Nevertheless, the interest of the scientific community in continuous computation arises now from several different perspectives, (Graca, 2004). Recent research in computing theory of stochastic analog networks (Siegelmann, 1999), neural networks and automata (Siegelmann, 1997, 2002) challenged the long-standing assumption that digital computers are more powerful than analog ones. The *analog* formulation of Turing's computability thesis suggests now that no possible abstract analog device can have more computational capabilities than neural networks, (Siegelmann & Fishman, 1998). In particular, the evolution of dissipative flow systems (as models for neural networks) can be interpreted as a process of computation where the attractor of the dynamics represents the output.

In (Siegelmann et al., 1999) the authors present a general theory of computation for dissipative dynamical systems. This theory interprets the evolution of such systems, both discrete and continuous in state space and in time, as a process of computation. Thereby, the notion of tractable (polynomial) computation in the Turing model is conjectured to correspond to computation with tractable (analytically solvable) dynamical systems having polynomial complexity.

**Field computation** (MacLennan, 1990, 1999, 2000) can be considered as a special case of neural computation which operates on data which is represented in *fields*. The latter are either spatially continuous arrays of continuous value, or discrete arrays of data (e.g. visual images) that are sufficiently large that they may be treated mathematically as though they are spatially continuous. A Fourier transform, e.g. of visual images, is an example of field computation.

This approach provides a good base for naturalistic computation. It is a model for information processing inside of cortical maps in the mammalian brain. Field computers can operate in discrete time, like conventional digital computers, or in continuous time like analog computers. Other realizations of field computation could be analog matrices of field programmable gate arrays (FPGAs) and grids thereof for image and signal processing problems. Further examples for field computation include optical computing (Stocker & Douglas, 1999; Woods, 2005), as well as Kirchhoff-Lukasiewicz machines (Mills, 1995) and very dense cellular automata (MacLennan, 2006).

Facing the pressure of Moore's law, much research on discrete quantum computers was directed towards the creation of smaller qubits that are subatomic in size. However, many quantum variables with *continuous* character, such as the position and momentum of electromagnetic fields, can be also useful. Noise is a difficult problem for discrete quantum computation, but continuous variables are more susceptible to noise than discrete ones. Thus, quantum computation over continuous variables becomes an interesting option towards a robust and fault-tolerant quantum computation and the simulation of continuous quantum systems such as quantum field theories. In (Lloyd & Braunstein, 1999), the authors provide the necessary and sufficient conditions for the construction of a universal quantum computer capable of performing "quantum floating point" computations for the amplitudes of the electromagnetic field.

**Biological computation.** Since the early $1950^{ies}$ discoveries of the structure of biopolymers (Pauling & Corey, 1951a/b, 1953) which defined the cornerstones of molecular genetics, there has been a long line of research towards decoding the information content of macromolecules. During the past 50 years computers have been intensively used to model and analyse complex organic structures. The age of *bioinformatics* began with the collection and analysis of large sets of data. Every new advancement revealing the contiguity of the underlying biological substances suggested new hypotheses, theories and methods for experimentation with them while paying tribute to their growing complexity.

The discovery of the double helix of DNA (Watson & Crick, 1953) which ascertained the *primary* structure of biopolymers in terms of a specific bonding among pairs of bases (adenine with thymine and guanine with cytosine) was followed by more elaborated research on the formation of macromolecules such as predicting the secondary RNA structure (Delisi & Crothers, 1971; Pipas & McMahon, 1975; Zuker, 1989; Washietl, 2000; Haynes et al., 2006). These morphogenetic studies were expanded towards exposing the biochemical and biophysical details of the tertiary and the quaternary structure of nucleic acids and proteins (Henrick, & Thornton, 1998; Klosterman et al., 2002; Tamura et al., 2004; Klosterman et al., 2004; Holbrook, 2005; Winter et al., 2006).

At the same time, complex biological structures such as those of nucleic acids have inspired research in biomolecular and biological computation. In 1987 Tom Head introduced a new method of relating formal language theory to studying the information content of macromolecules (Head, 1987). He associated a language with each pair of sets such that the first set consists of double stranded DNA molecules and the second one corresponds to recombinatorial behaviours allowed by specific classes of enzymatic activities.

This language consisted of (linear) *strings* of symbols representing the primary structures of the DNA molecules that may potentially arise from the original set of DNA under given enzymatic activities. The result was analysed then by means of novel generative *splicing* system formalisms. In the following decade the theoretical base of biomolecular computation was developed.

In 1994 Leonard Adleman used a set of simple DNA operations to solve an NP-complete problem as a 7 node instance of the Directed Hamiltonian Path (Adleman, 1994). The performance of this device outperformed by more than a million times the one of the fastest supercomputers at that time. This work set out the beginning of a new era of scientific exploration at the edge between biology and mathematics, known as *DNA computation* which became an instance of *biocomputation,* (Kari, 1997). A comprehensive multi-disciplinary survey of biocomputing is given in (Hong, 2005a/b).

The following decade was devoted to the study of computation principles such as bounded nondeterminism and efficient recursion (Beigel & Fu, 1997) and to the further development of formalisms from insertion-deletion and equality checking systems (Zingel, 2000) to graph transformations (Harju et al., 2002), DNA word concatenations (Andronescu, et al., 2003) and circular gene splicing languages (Bonizonni et al., 2005). The exploitation of massive parallelism in biomolecular structures was a key element of this research wave (Reif, 1998-1999).

Furthermore, relations to other scientific fields such as control and chaos theory were investigated (Manganaro & de Gyvez, 1997). During this period, experimental studies of the mechanisms for the manipulation of biomolecules, such as kinetic partitioning (Thirumalai et al., 1997), were also pushed forward. Biomolecular computation ventured into the domains of game theory, neuroscience and design automation to investigate utilization aspects such as RNA solutions to chess problems (Faulhammer et al., 2000), experimental DNA neural computation (Mills et al., 2001) and bottom-up circuit patterning based on DNA self-assembly (Dwyer, 2005). Other prominent developments were autonomous DNA nanomechanical computation and motion devices (Yin et al., 2005), as well as autonomous DNA cellular automata (Yin et al., 2006). The challenges and applications for self-assembled DNA nanostructures were thoroughly reviewed in (Reif et al., 2001). A recent survey of this topic is given in (Reif & LaBean, 2007). Qualities such as powerful parallelization, small size with an enormous storage capacity and efficient energy utilization are now driving this research towards evolutionary biological, biomolecular and organic computing (Salthe, 1985; Eberbach, 2005; Henkel, & Joost, 2005; Stadler, 2007). A comprehensive overview of computing in living cells is given in (Ehrenfeucht et al., 2004).

Considering the above, and the layered structure of biopolymers such as nucleic acids and proteins which is supposed to deliver an abundance of possible spatial relationships between and within the composing elements (strands, bases, atoms, etc.) at different levels and hence to involve *multiple operational semantics* depending on the specific context of the interactions among the structural elements and their situational environment, it is remarkable that such unilateral and oversimplified models are suggested for biocomputation and are often taken for granted in bioinformatics. The train of thought in this field continues to follow the standard deterministically-combinatorial Turing Machine model. However, a DNA molecule can have at least three layers of formal symbolical representation in the 2D plane. It can be thought of as a string over an alphabet of:
- 20 symbols, each one representing a specific amino acid;
- 4 symbols, each one representing a rhibonucleotide, distinguished by the chemical group (base) attached to them (*adenine*: A, *guanine*: G, *cytosine*: C, and *thymine*: T) or
- 4 symbols, each one representing a hydrogen-bonded deoxyribonucleotide pair.

Therefore, it can semantically encode at least three different modes of discrete computation which practically occur simultaneously. Thus, a single structural change within a DNA strand can be explained by three different computation models at the same time. Selecting only one of them can be reasonable and useful in the special controlled conditions of artificial life systems, but they all remain hypotheses in *real life* situations, unless they are proved by experiment in all their aspects.

To identify and extract patterns responsible for the latent semantic and semiotic encodings within multilateral relations among the basic structural elements and their higher level organizations it is necessary to analyse macromolecular structures based on the data about their coherency in terms of their biochemical (Pauling & Corey, 1951a/b, 1953: Erdman et al., 1983; Holbrook, 2005) and biophysical (Gariaev, 1994, 1997) properties. However, it must be recognised that true encoding, storage (inheritance) and biocomputation involves not only DNA nucleotides, but also protein sequences, non-protein coding RNAs, mitochondria, microtubules, lipid membranes, nuclei, cells, endoplasm, organs and the entire organism (Noble, 2006). The whole process is tuned by genetic buffering the restrains of present and past interactions at each level with the surrounding environment (Venter, 2007). In other words, "the program is the system itself" (Noble, 2008).

Which model is the correct one? Perhaps there are more models or none of them are (particularly) correct? How do we know what a DNA or RNA macromolecule "keeps in mind" when folding this or that way? Which detail of the situation is essential? What is a gene indeed? Is it a distinct physical entity like the electron or a virtual one manifested in the context of a specific situation or relationship? If the second option is true, what is a gene in terms of physical elements and their multiple structural-functional relationships? Which role has energetic valence in structural formations?

When, how and why does a new gene emerge? How does it change and evolve? These are the questions that should be answered by truly genetic and naturalistic programming.

The above review of alternative models to classical computation is not exhaustive, but it provides a good starting base for the transition to the next section which introduces the main part of this contribution.

## 5. Integral Biomathics

Although being now a special case in modern physics, Newton's world picture still dominates computer science (Blum, 2004), engineering (Lee et al., 2005) and biology (Bower, 2005). Turing Machines are Newtonian in the broader sense that they deal exclusively with syntax and inference rules based on discrete logic in absolute space and time to deliver predictable behaviour.

Yet, the review of the diverse non-classical computation approaches beyond the Turing frame in the previous section leads to the conclusion that these models hypothesize different post-Newtonian laws of physics as a precondition for their implementation.

Although modern computer science has not really entered the relativistic sub-nuclear age of modern physics yet, the abundance of computational ideas and approaches indicates that researchers are well aware of the limitations of the Turing computation model. They are certainly going to use every discovery and invention in physics to realize their concepts. Perhaps the most significant aspect of this finding from a historical perspective is the comeback of analog computing now based not on mechanical components and electronic circuits, but on artificial neural networks and continuous quantum computation. This is a very interesting fact which shows that computation is now closer to biology than to classical physics. Indeed, computer science and biology maintain today a similar relationship to that of $19^{th}$ and $20^{th}$ century mathematics and physics. Progress in one field will influence progress in the other and vice versa, for it is not possible to avoid *analogy* and *semantics* (in terms of the formal logic known today).

The author cannot limit himself to a priori established conventions (based on past facts) about what is general and what is special in theoretical research. In this respect he is led by the ideas presented in the introductory chapters of Rosen's *Life Itself* (1991).

From the discussion in the previous two sections it becomes evident that the two fields or *planes* of research are closely interdependent through synergy and correlation and by addressing a number of parallelized phenomena, paradoxes and questions. Thus they represent a congruent pair, a dyad, of knowledge in complex system design and automation that deserves special attention from the scientific community. It is therefore the author's intention to get beyond the frames of autopoietic theory and Turing computation and explore new computational models in complex living autonomous systems. His focus represents the research in natural automation and computation, including models of information emergence, genetic encoding and transformations into molecular and organic structures.

This area is the joint domain of life sciences, physical sciences and cybernetics. In this respect, the author is interested in investigating natural systems which are robust, fault-tolerant and share the characteristics of both living organisms and machines and which can be implemented and maintained as autonomous organisations on molecular and atomic scale without being planned. Therefore, he is going to use concepts, models and methods from such disciplines as evolutionary biology, synthetic microbiology, molecular nanotechnology, neuroscience, quantum information processing and field theory.

The author proposes to enhance these concepts with other techniques and formalisms from classical and non-classical computation, network and information theory to address specific challenges in the pervasive *cyberzoic* era of integral research beyond nano-robotics and nano-computation into *post-evolutionary* computation, swarm intelligence, adaptive behaviour and co-evolution of biosynthetic formations, molecular self-assembly, synthetic morphogenesis and evolvable morphware (Zomaya, 2006). What is meant by this is ultimately the convergence and transformation of biology, mathematics and informatics into a new naturalistic discipline the author calls **Integral Biomathics**.

This new field is founded on principles of *mathematical biophysics* (Rashevsky, 1948; Rosen, 1958), *systems biology* (Wolkenhauer, 2001; Alon, 2007) and *information theory* (Shannon, 1948) endorsed by biological communication and consciousness studies such as those described in (Sheldrake, 1981; Crick, 1994; Penrose, 1996; Hameroff, 1998; Edelman & Tononi, 2000). Thus "…whatever is in a part must be in the whole..." (Lodge, 1905, p. 64)

Here the term 'integral' implies also 'relational' and denotes the associative and comparative character of the field from the perspective of *cybernetics* (Bateson 1972) and *systemics* (François, 1999).

Finally, Integral Biomathics embraces *internalist* concepts and theories aiming at describing vague, impermanent, unbalanced and generative domains such as Uexküll's Umwelt (1909, 1937), Bohm's internal variables (1952; 1995), the autopoiesis of Maturana and Varela (1980), Salthe's evolving hierarchies (1985), Matsuno's protobiology (1984/1989) and Rössler's & Kampis' endophysics (Rössler, 1987/1998; Kampis, 1993/1994). While focused only on local variables in the course of system changes, the internalist approach attempts to understand a system from within, with the inquirer being a part, inside the system. It is complementary to the traditional externalist scientific models. Therefore, the ultimate goal of Integral Biomathics is the consolidation of the numerous system-theoretic and pragmatic approaches and technologies to artificial life and natural computation and communication within a common research framework. The latter pursues the creation of a stimulating ecology of disciplines for studies in life sciences and computation oriented towards naturalistic system engineering where "the program *is* the system itself", (Noble, 2008).

The author's approach does not antagonize old theories and results. It does not defeat recent or new ones either. In this way, the author answers the rising appeal by systems biologists to develop integrative and reconciling philosophies to the diverse approaches to genetics, molecular and evolutionary biology, (O'Maley & Dupré, 2005). Thus, following the line of research set up by the pioneering works of Rashevsky (1940 ff.), Turing (1952), Rosen (1958 ff.), Bateson (1972), Miller (1978), Maturana and Varela (1980)**,** Integral Biomathics will address questions arising in the widened relational theory of natural and artificial systems.

This discipline is concerned with the *evolutionary dynamics of living systems* (Nowak, 2006) in a unified manner while accentuating the higher-layer dynamic relationship, interplay and cross-fertilization among the constituent research areas. In particular, Integral Biomathics is dedicated to the construction of general theoretical formalisms related to all aspects of emergence, self-assembly, self-organisation and self-regulation of neural, molecular and atomic structures and processes of living organisms, as well as the implementation of these concepts within specific experimental systems such as in silico architectures, embryonic cell cultures, wetware components (artificial organic brains, neurocomputers) and biosynthetic tissues, materials and nano-organisms (cyborgs).

Therefore, the research methods of Integral Biomathics include not only the traditional ones of experimental and theoretical biology, involving such disciplines as molecular biology and functional genomics, but also the *dynamic inclusion* of novel computational analysis and synthesis techniques which are characteristic for the corresponding frame of relevance, FoR (MacLennan, 2004) and beyond the existing taxonomic framework for modelling schemes (Finkelstein et al., 2004). This corresponds to a qualitatively new development stage in systems biology and engineering. The algorithmisation of sciences not only placed biology closer to the traditional 'hard' sciences such as physics and chemistry, but also provided the base for a paradigm shift in the role distribution between biology and mathematics, (Easton, 2006). Therefore, the goal of this new field – biologically driven mathematics and informatics (*biomathics)*, and *biological information theory* – is the elaboration of naturalistic foundations for synthetic biology, systems bioengineering, biocomputation and biocommunication which are based on understanding the patterns and mechanisms for emergence and development of living formations.

The author's major objection, however, to previous efforts for unification in this field relates to the roles of *causation* and *entailment* in the process of creating and organising life forms. These processes are *qualitatively different* from the system models known in physical sciences. Therefore, Integral Biomathics aims at: (i) removing the restrictive reductionist hypotheses of contemporary physics, (ii) adopting the appropriate mathematical formalisms and models, and (iii) deriving new formalisms out of the biological reality that faces the complexity of living systems more adequately. Such arguments are in line with pioneering research in mathematical biophysics (Rashevsky, 1948, 1954) as well as with recent results in systems biology (Westerhoff et al., 2004; Mesarovic et al., 2004).

Therefore, the author provides an extended model about the interdependence between the various disciplines in this field based on Rosen's category-theoretical definition (Rosen, 1991, p. 60).

Figure 1 illustrates Rosen's relational model of science (associated with physical systems) based on the concept of *Natural Law* (Whitehead, 1929, 1933). The latter represents a bidirectional mapping between the Natural World and the Formal World which is depicted by the double edged large arrow in the middle of the figure. The two brain images with the prisms on them below and above this arrow symbolically break the light from a source world to a target world by means of human perception in two opposite directions.

This model relates a mechanistic world of conservative closed systems (known e.g. from Newton's laws of motion and celestial mechanics) to a discrete logic world of state automata and algorithms and vice versa. It represents the most common model of natural science and engineering. The arrows 1 and 3 represent the recursive entailment structures within the corresponding domains – causation in the natural (physical) world and inference in the formal (abstract) world – whereas the arrows 2 and 4 express the possibility of consistent use of syntactic and semantic truth through encoding/measurement (Kuhn, 1961; Rosen, 1978) and decoding/realization. In particular, arrow 2 depicts the flow path of *abstraction* and generating hypotheses about the natural world in physical (analytical) sciences. Arrow 4, in turn, represents the path of *de-abstraction* and creating forecasts about natural world events using formal models and theories.

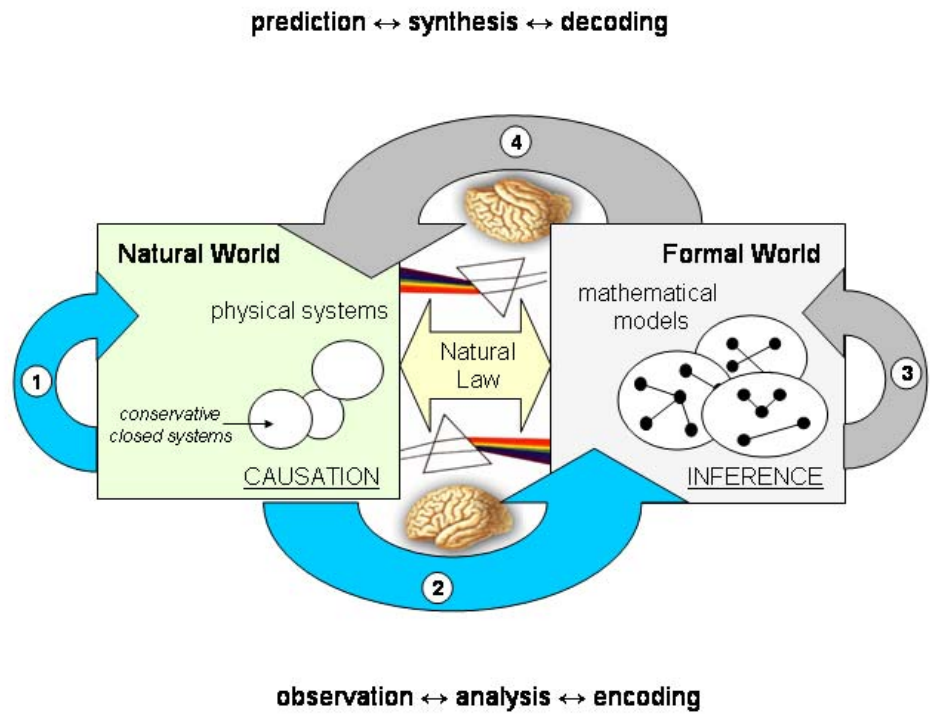

**Figure 1**: Rosen's modelling relation of science (Rosen, 1991, p. 60)

These forecasts are then used to prove the truth of scientific theories, inferred from hypotheses in the formal world, by observation and measurement in the "internal" loop 1 of the physical world. Hence, there are two separate paths of cognition in (physical) science: 1 and 2 + 3 and 4. If both of them deliver the same replicable results, the theory derived from the hypothesis in the formal world becomes a *model* of the natural world, (Rosen, 1991). At the same time, arrow 4 shows the path of invention and engineering of artificial systems in synthetic sciences (e.g. computer science) that emerge out of mathematical models and theorems within the "internal" loop 3 in the formal world. Thus, arrow 4 can be regarded as a double one.

In engineering, when one starts with inference systems in the formal world (e.g. human minds) that are intended to realize physical world systems systematically, i.e. "by natural law", one faces the so-called *realization problem,* according to Rosen, which involves "modes of entailment falling completely outside contemporary science" (e. g. knowledge through intuition mentioned in section 3).

Indeed, both arrows 2 and 4 of encoding and decoding remain *unentailed* according to Rosen's model. The reason is that there is no mechanism *within* the formal world to change an axiom or a production rule and there is no such mechanism within the physical world to change the flow of causation. These arrows are neither part of the natural world, nor of its environment. They do not belong to the formal world either. They appear like mappings, but they are not such in any formal sense.

This finding is consistent with Gödel's incompleteness theorems[2] that state there is always a true statement within a formal system that cannot be proved within that system and requires a higher level formalism.

---

[2] In particular, the second incompleteness theorem: "*For any formal theory T including basic arithmetical truths and also certain truths about formal provability, T includes a statement of its own consistency if and only if T is inconsistent.*" (Gödel, 1931; Bagaria, 2003)

The above relationship model holds for describing *closed conservative systems* in a Newtonian world based on causation which has its counterpart as a formal world model based on a set of axioms and inference rules which are ultimately represented by Turing's concept of state automaton. A better representation of reality, however, would also account for dissipative open systems, Fig. 2.

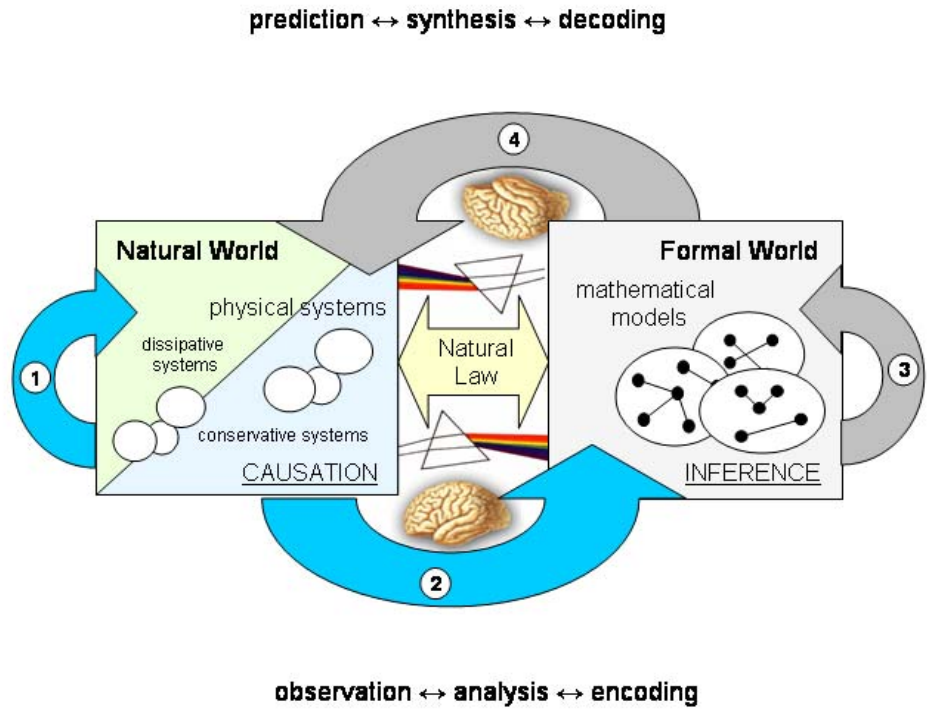

**Figure 2**: The extended modelling relation of science including dissipative systems

[Some authors like Salthe (2007, pers. corresp.) take the view that living systems were originally abiotic dissipative structures that either were 'invaded' (whatever that may mean!) by the genetic apparatus or that the genetic apparatus emerged within some kind of dissipative structure.] This graphic symbolically introduces asymmetry and self-organization in the model. A fundamental theorem of mechanics proved by Emmy Noether states that symmetry gives rise to conservation laws that inhibit self-organization; hence *asymmetry stimulates self-organisation*, (Noether, 1918).

Here, the physical world on the left-hand side of the figure is split into two parts, the one of closed conservative systems governed by causation, and the one of dissipative systems, where both chaos and order, ergo self-organization, are at work. Whereas the first field provides the physical source and target for formal computation and automation, the latter has been of little or no concern in science for the purpose of creating artefacts and machines. This is one of the central issues this paper points out.

In a further step, the author adopts Rosen's postulate about the generalization of biology over physics (Rosen, 1991, 1999) and requires that physical systems, conservative and dissipative ones, interacting with or exchanged between biological systems be inclusive within their organizational closure (Nomura, 1997 ff.; McMullin, 2000b).

In other words, *specific physical systems within biological context* (resources) *are declared as a subset of the more complex set of biological systems* (and not the reverse). This is in fact a model extension and not a generalized assumption.

The concept is similar to reversing the reductionist bottom-up causal chain in biology {genes → proteins → pathways → cells → tissues →organs → organisms} into a kind of "downward causation" (Noble, 2006, 2008) in the opposite direction (from the complex to the simple) within the context of living. Of course, this view is also incomplete, but it provides a good focal base for understanding the train of thought advocated further in this paper.

Indeed, there is no contradiction with the generally adopted antithesis that a physical system cannot be logically, or an example of, a biological system. This means that the model in figure 3 does not disagree with the hierarchical relations (Salthe, 1985) between the realms of Nature. It only states that the physical realm is more restricted than the biological one within the organizational closure of the latter. In other words, the model restates the Biosphere/Noosphere/Gaia concept (Vernadsky, 1945, 1998; Lovelock, 1979).

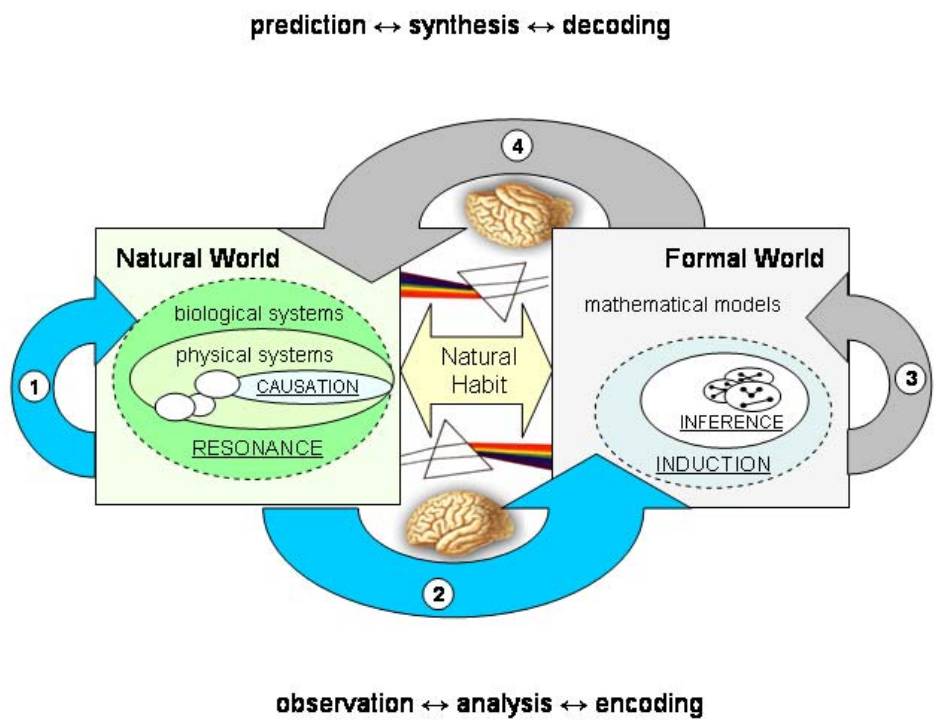

**Figure 3:** The revised modelling relation of science with generalization of biology

In addition, the author proposes three substantial amendments to Rosen's modelling relation defining the major distinctions of biological systems when compared to physical systems. Figure 3 depicts these changes which extrapolate and endorse the interchange circle model between synthetic biology and systems biology presented in (Barret et al., 2006).

This relationship is also analogous to Salthe's thermodynamic cycle model of energy gradient degradation catalyzed by form (Salthe, 2002), illustrated as formalization.

Here, the author adopts and develops some concepts related to the *morphogenetic field* (Gurwitsch, 1910 ff.; Weiss, 1939) or *inductive force field* (Faraday, 1873) in evolutionary and developmental biology (Thompson, 1917; Franck, 1949) and the *formative causation* or *morphic resonance* hypothesis (Sheldrake, 1981; cf. section 6 for more details), as basic organization principles in biological systems, analogous to those in quantum physics. His motivation for addressing this relationship is that morphogenetic oscillations are at the base of i) *pattern formation* and ii) *conservation* in living systems while being closely related to iii) *energy*.

The following examples advance the above argument

i.   During regeneration, the cell ball of the fresh-water polyp *Hydra vulgaris* undergoes subsequent shape transformations with *symmetry* breaking (Noether, 1918) being accompanied by characteristic oscillations in size and shape (Fütterer et al., 2003).

ii.  The possibility of *self-sustained oscillations* of chemical concentrations during morphogenetic processes was investigated in (Granero-Porati & Porati, 1984). The authors have shown that in the case of the Gierer-Meinhardt model for biological pattern formation (Gierer & Meinhardt, 1972), self-sustained oscillations are possible when an order parameter, connected with the decay constants of morphogenes, crosses a critical value. The obtained analytical results were in agreement with the numerical results of Gierer and Meinhardt.

iii. Single amoebae aggregate when their food supply is exhausted. This is the result of a succession of periodic coordinated movement steps directed towards a group of amoebae — the "centre", which produces cyclic AMP (the signal) in an oscillatory manner, whereas other cells in the field respond chemotactically by moving up cyclic AMP gradients and relay the signal progressively outwards. Experiments have shown that the oscillation frequency increases with temperature (Nanjundiah et al., 1976).

Further, field interactions and vibrations are also characteristic in string theory. Hence, "the universe – being composed of an enormous number of these *vibrating strings* – is akin to a cosmic symphony" (Greene, 1999).

Thus, the principle of resonance does not exclude causation, but rather integrates it within a general frame as in the case of electromagnetism (Faraday, 1844-1855; Maxwell, 1865; Hertz, 1887; Tesla, 1892a). In this model the author replaces the human-centric concept of 'natural law' by the biological one of *natural habit* (coined by Sheldrake). The term "law" is too strong for biology, whereas habits are less restrictive towards changes. Natural habit is therefore synonymous with 'natural pattern'. Finally, the author applies the principle of 'lateral' *induction* in the formal world for creating new formalisms about natural systems through pattern recognition and *analogy*, based on observations, experiments and *associations* with other formalisms.

Note, the term "lateral" is used to discern from the classical Peircean definition of induction in semiotics, namely the generalization, the implication of a sign from a specific (syntactic) set of properties of an external (vague) object that have been acquired randomly a priori through development, measurement and refinement by the observer (Salthe, 1998). Lateral induction (or simply induction in Integral Biomathics) is related to the abductive construction of a new concept, model, or category, yet it can also have no traceable propositions in the logical sense, but rather complete assumptions and hypotheses that emerge spontaneously as if being induced or inspired by an external mind.

This special definition of induction reflects the true nature of invention, artistic creation and discovery, thus corresponding to de Bono's concept of 'lateral thinking' (de Bono, 1967); cf. section 6 for more details. It is analogous to morphic resonance in the natural world and addresses the larger set of formal systems that reflect the behaviour of biological systems from the standpoint of Integral Biomathics. In this way, biology is generalized and physics becomes its special case with causation and inference being entailments of resonance and induction respectively.

Note that the external circles of biological systems and their respective formalisms on both sides of the graphics are dotted to represent open, *living* systems. Furthermore, in the bidirectional relation of *natural habit* between the two worlds in figure 3, the reader can identify two epistemological processes denoted by arrows 2 and 4.

Firstly, there is the process of pattern recognition in the natural world and the phenomenon of memory viewed by the self or the formal world (arrow 2). Secondly, there is the case of *relational science* in the formal world based on formal theories of inference in the first concentric circle of mathematical models, but also on metaphors, analogies and non-local relations or induction in the second incorporating circle (arrows 3 and 4).

It is still not known how an idea emerges from, relates to, coincides/reacts with or is influenced by other ideas in human minds. It is theorized, however, that thoughts and consciousness exhibit such properties as *synchronicity* (Sinclair, 1930; Jung, 1952; Jung & Pauli, 1952)*, non-locality* (Bell, 1964; Targ & Puthoff, 1978), *quantum entanglement* (Aspect et al., 1982; Donald, 1991; Stapp, 1993), *quantum coherence* (Penrose, 1996; Hameroff, 1998) and *bounded autonomy of levels* (Mesarovic & Sreenath, 2006). So far, science does not know the idiosyncratic nature of such mechanisms as (spontaneous) inversion, correlation, fusion and fission of concepts. This is an interesting research area along the pathways of arrows 3 and 4. Arrow 4 is also associated with the next level entailments in the formal world and with *resonance* in the natural world related to such phenomena as life, thought and consciousness. Thus, arrows 2 and 4 represent an open infinite helix/*spiral* of scientific development rather than a closed stuttering loop.

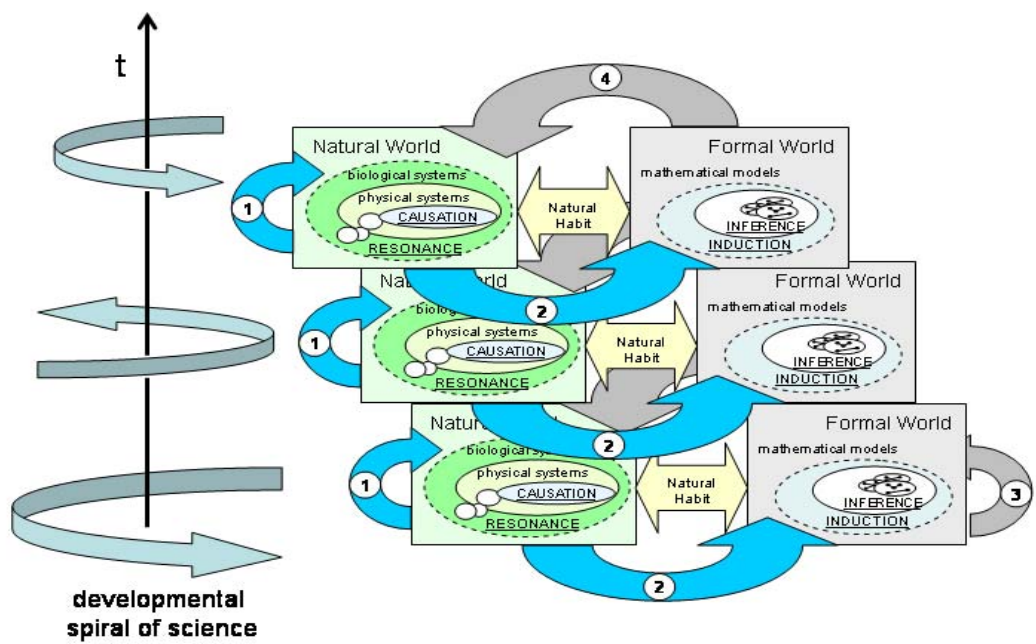

**Figure 4**: The evolving modelling relation for science

Figure 4 represents the evolution of formalizations and realizations with the shifted perspective of higher layer entailments along the time axis. It suggests that a vertical view of the slices of shifted flat world layers along the temporal axis may deliver new insights into the cross-layer links between the entities in the diagram and the real dimensions of the interplay within the formal and natural worlds.

This layered interrelationship within the developmental spiral of the world "versions" results from paradigm changes and inflection points in the formal world. Hence, the uplifting force in this development process is represented by arrow 4 at all layers in figure 4, which is the human mind, a living system and part of the Natural World. The goal of this driving force is the fusion of the two worlds, the natural one and its mental representation along the line of creative change. Hence, we can ultimately speak of a United World. Yet, it cannot be perceived and developed without the separation of concepts expressed in figures 1-4, (Kuhn, 1961-62).

Complexity in the natural world is manifested through implicit and explicit order. The implicit order can be encoded in 'hidden variables' (Bohm, 1952) that enable semantic enfolding and unfolding in the formal world (Bohm, 1980; 1995). The later is associated in the author's enhanced model with recursive pattern generation (including failures) followed by reflexive processes of evaluation, change and adaptation in response to external stimuli in the presence of noise and disturbance.

Essentially, the author proposes an integral evolving model of layered dynamic interdependence between the logical formal meta-computational reasoning system world and the natural autonomous biophysical system world. The two worlds represent a dyad in perpetual development which ultimately embeds the unentailed relationships of Rosen's original modelling relation shown in arrows 2 and 4 in figure 1.

This process is analogous to Werner's claim that "ultimately, *in silico* artificial genomes and *in vivo* natural genomes will translate into each other, providing both the possibility of forward and reverse engineering of natural genomes", (Werner, 2005). In fact, the evolution of the interpenetration of the natural and formal worlds is redolent of the Biosphere/Noosphere/Gaia idea (Vernadsky, 1945, 1998; de Chardin, 1955; Lovelock, 1965; Lapo, 2001) which is "probably the tendency on the surface of the earth" (Salthe, 2007, pers. corresp.).

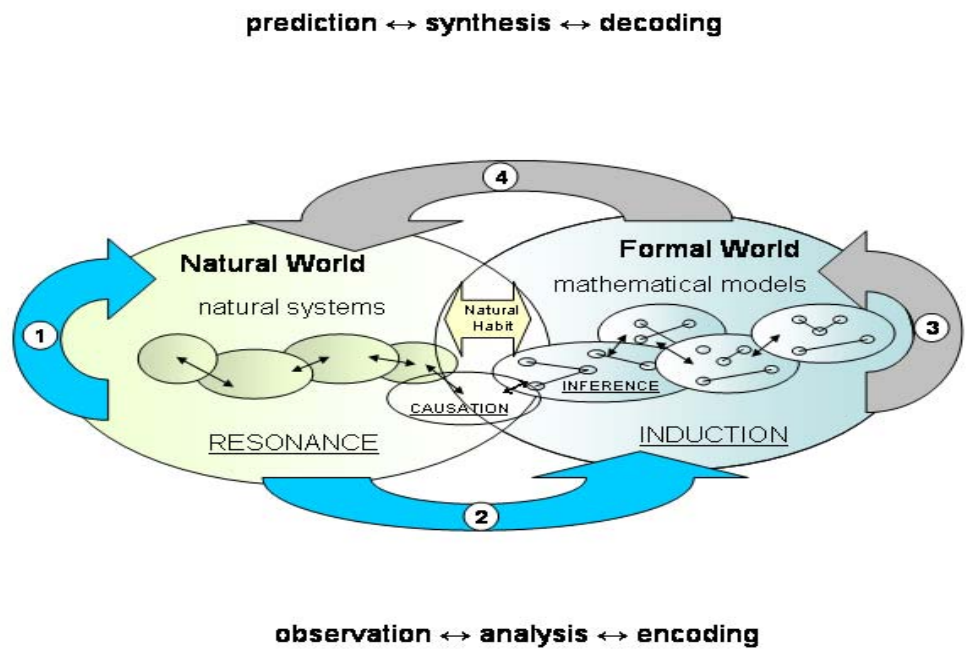

**Figure 5**: The converging model relation for science

Figure 5 illustrates the result of this process which expands the author's model of *Wandering Logic Intelligence* (Simeonov, 1999-2002a/b/c) to science. Causation and inference may continue having their restricted domains of relevance. However, resonance and induction will be recognized as essential.

The above amendments in Rosen's modelling relation are necessary because *in a historical perspective mathematics has been derived from and developed for descriptions of physical phenomena.* Good examples of this are: (i) Pythagoras' monochord, the principles of harmonic tonal mediation and the formation of ratios, (ii) Leibniz's mechanics method for the retrieval of infinite proportional rows and logarithms (Leibniz, 1691), and (iii) Newton's fluxion calculation which became the base for the integral calculus (Newton, 1686; 1693/1704).

Hence, there is a *'hidden' semantics in mathematics* that has proven to be a suitable tool for studying static mechanics, celestial mechanics, thermodynamics, electromagnetism and relativistic theory. However, most of its formalisms are insufficient to handle biology in a straightforward manner. Therefore, it is necessary to develop new mathematical foundations for modelling different types of biological systems just as von Neumann and Dirac proposed their own formalisms to explain quantum mechanics.

Indeed, the phenomenology of quantum theory is closely related to that of theoretical biology (Rashevsky, 1961a), which means that biomathics could successfully adopt and develop the formal toolset of quantum theory. It should be kept in mind, however, that "observations in QM are all carefully engineered phenomena, while observations of biology are directly phenomenal" (Salthe, 2009, pers. corresp.). It will not be easy to bridge both ends within such a biophysical formalism.

In view of these new developments, it is also very likely that the understanding and the definitions of *mathematics, computation* and *cognition* in nature and engineering will be revised. The basic characteristic of physical systems is that they are closed and that everything that could not be included within a closed system is neglected. However, in biology the reader faces an inverted world to that of physics, because systems are basically open (Uexküll, 1909; 1912; 1920a; 1937) and because the second law of thermodynamics about the system steady state is defined in terms of order or *negentropy* (Schrödinger, 1944; Schneider & Kay, 1994; Gladyschev, 1997), instead of chaos and entropy.

Therefore, there is a need for a different kind of mathematics to support a new kind of science that is devoted to the discovery of *recursive patterns of organization* in biological systems (Turing, 1952; Gierer & Meinhardt, 1972; Miller, 1978), such as those of neural activity where neural systems could be understood in terms of pattern computation and abstract communication systems theory (Andras, 2005) or unicellular chaos dynamics (Aono et al., 2007).

This science should be able to deal with the complexity of biological systems by restructuring its ontology base to correct its models whenever appropriate. At its foundations it should be capable to express and retrace: i) *resonant causality* "by which several physically separated and highly selected areas [of an organ(ism)] may simultaneously be affected because of a shared and underlying *wave phenomenon* in some medium" [italics by the author, cf. (Gariaev, 1994, 1997)], and ii) adaptive and anticipatory control, i.e. "proper timing in *continual* (i.e., periodic sampling) *feedforward relation to the future*" (Musès,1985; pp. 49, 58).

Figure 6 illustrates the ultimate evolving model of science where the natural and the formal worlds increasingly interpenetrate to finally form a monad, the unity of both realities comprising artificial and natural systems. Here, the arrows 2 and 4 (Fig. 1-5) disappear for they are now incorporated within the growing internalist mapping (Matsuno, 1984/1989; Rössler, 1987/1998; Kampis, 1993/1994), the double arrow of *Natural Habit*; hence arrow 3 becomes arrow 2 for consistency.

The formal world represents a generalization of all models and not only the mathematical ones. It is essential, that the distinction between the formal and the natural world decreases in time, as well as the one between their driving forces, **induction** and *resonance*. Whereas induction triggers the "critical mass" of information to invoke synchronicity in the emergence of ideas, models and projects in different and often distant fields of knowledge, resonance reflects this connection in physical domain. In other words, the evolution of the two worlds, Mind and Matter, will lead to a convergent reality.

The same figure demonstrates the layering of non-classical models of computation discussed in section 4, as well as the evolution of science associated with the latest cosmological hypotheses about the gradual change of physical laws (Smolin, 2003).

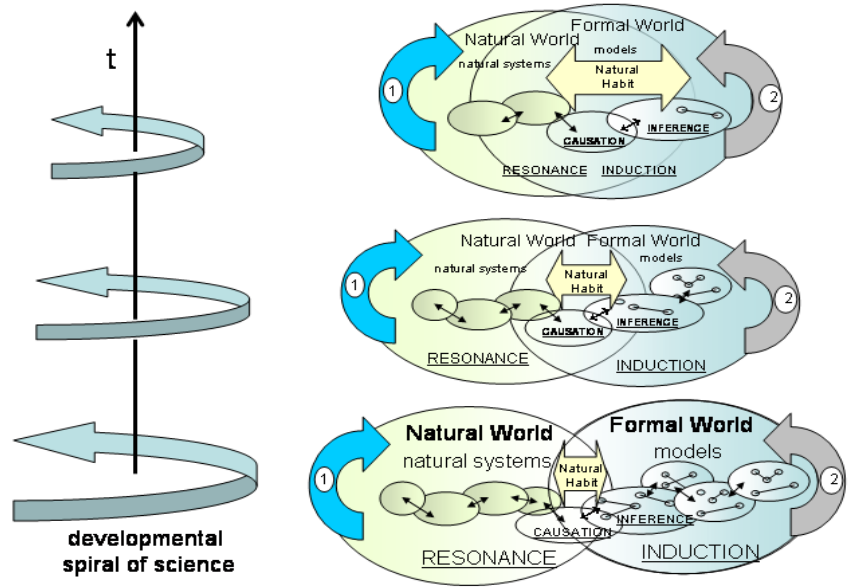

**Figure 6**: The ultimate evolving modelling relation for science

Ultimately, this development suggests one natural world that has been once split into an external and an internal part which are then unified again by the human mind after having induced some changes in both spheres. Furthermore, it can be hypothesized that when the point of singularity is reached, one single state of *resonance* (accord) upon negentropic structures and processes will dominate over this world until the collection of critical information mass initiates another transformation in the opposite direction (towards growing entropy) or in another dimension that we cannot predict yet.

It can be speculated that some of the innate mental processes of arrows 3 and 4 on Fig. 1-5 can be recognized and associated with the orthogonal plane of subjective perception, direct knowledge awareness and unbound understanding or *satori* (not shown on the figures). This is indeed an old issue, for the "mystic" experience of *intuition, insight* or *inspiration* (muse) is indisputable in art and music. The western Cartesian tradition of science has been ignoring this "detail" for four centuries.

## 6. Discussion

The approach to Integral Biomathics presented in the previous section appears related to the classical dialectical scheme of thesis-antithesis-synthesis, (Hegel, 1807). Yet, the author has a different motivation for explaining the philosophy of this new discipline that was derived from the analysis of the latest developments in the participating scientific fields discussed in sections 1-4.

It is important to apprehend how previous concepts and theories in science influence our present day lives and futures. Therefore, the new science which is about to emerge has to be open, liberal and *response*-able (responsible), one that abides and unites anomalies, paradoxes and contradictions within its frame. It should not ignore, deny, combat or control the controversial theories at any price, but rather let them co-exist and complement each other into a semantically rich, interpretation of reality.

In particular, the author identified the following categories that play special roles in Integral Biomathics:

**Syntax vs. Semantics and Semiotics.** Formal concepts are either purely syntactic in Hilbert's sense or they are created through recursive extrapolation from other formal concepts. In the second case, they contain a semantic component of truth that links them to the context of those previous concepts they were derived from. Rashevsky pointed out that there are different ways in mathematics to approach the relational problem in biology using such means as set theory, topology or group theory (1961). Each one of these approaches is able to represent different aspects of relations in a system within different contexts. Thus, context (semantics) and relation (syntax) can be interchanged depending on the purpose of the description (Chomsky, 1953-1956; Minsky, 1968; Fodor, 1987). Turing computation and the mathematics of computation are essentially syntactical (since they are devoid of semantics), but the computation of tactics in a biological environment could be argued to be implicitly semantic.

Therefore, the author regards even pure syntactic models in the formal world as semantic and semiotic inclusions of axiomatic truths which are obtained in an empirical way from the physical world through the associative link of natural law/habit. A close example for this fact is Newton's derivation of the rule for integration by comparing 2D surfaces traversed by the movement of points firmly attached to physical bodies (Newton, 1693, 1704). In order to change the historical discrimination between *syntax* and *semantics* in the formal world, (Shannon, 1948), imposed by the different perspective of the present day understanding of the natural world, it is necessary to step back and generalize the description of the domain of discourse. This can be realized by taking out the restrictions placed by Newtonian mechanics and the Turing Machine model to a degree that allows us to reconcile traditional Turing based (incl. Super-Turing) and non-Turing based approaches in computation (Rosen, 1991; Hogarth, 1994).

**Dissipative vs. Conservative Systems.** Life belongs to the class of natural phenomena which are not isolated and simplistic, but related and complex. Therefore, modelling computation and cognition in living systems is analogous to studying non-equilibrium / dissipative open systems capable of exchanging information in the context of noise, disturbances and fluctuations of their environment.

**Resonance vs. Causation.** Sheldrake's hypothesis of formative causation or morphic resonance states that morphogenetic fields shape and organize systems at all levels of complexity (atoms, molecules, crystals, cells, tissues, organs, organisms, societies, ecosystems, planetary systems, galaxies, etc.). Accordingly, "morphogenetic fields play a *causal* role in the development and maintenance of the forms of systems", (Sheldrake, 1981, p. 71); they contain an inherent memory given by the processes of morphic resonance in the past, where each entity has access to a collective memory. The presented approach differs from Sheldrake´s original definition in two points. Firstly, the author palpably distinguishes between *causation* and *resonance* as universal organization principles of the Natural World (Fig. 2). Whereas in causation multiple linear cause–implication chains about events in the domain of discourse can be clearly identified, he understands resonance usually as a non-linear, non-local and (sometimes) semantically 'hidden' spatio-temporal relationship between entities in the broad sense. The entailment of causation within resonance is however allowed in the author's model analogously to the scales of interactions in physics. Secondly, the author regards resonance as a dyad consisting of (i) *energetic* resonance as it is known in physics (wave mechanics, electromagnetism, quantum mechanics, string theory, etc.), and (ii) *information* resonance corresponding to both classical communication theory (Shannon, 1948) and to Sheldrake´s morphogenetic field theory.

**Induction vs. Inference.** By the term 'induction' in Integral Biomathics the author does not mean the classical mathematical induction used for formal proofs. It is the counterpart of resonance within the formal world which entails not only the classical formal reasoning theories in mathematics, but also (yet) unknown structures and pathways of logic based on complex, semantically enfolded relationships within and between the formalisms as "musico-logical offerings" (Hofstadter, 1999). The reader can imagine induction as the self-organized process of generating and evolving formalisms from a set of basic axioms and theorems that can *mutate* depending on the results of cognition.

**Synchronicity and Choice vs. Determinism.** When the reader observes the above categories as an evolving dynamic set of cross-interacting patterns of structures and processes for organization and exchange between the natural and the formal worlds, she can realize the third epistemological dimension about the multi-relational helix gestalt of scientific exploration along the temporal axis (Fig. 3). The author believes that the usage and verification of this macro-model can deliver new insights into the natural phenomena of synchronicity and choice in the context of emergence, differentiation, organization and development of biological structures and processes.

Following the line of thought from the previous paragraphs, the author hypothesizes that determinism is entailed within synchronicity and choice. The above considerations demand a systematic study of the traditional and the opportunistic approaches to computation and automation along with their relationships and frames of relevance.

## 7. Implications

Recent efforts in autonomic computing (IBM, 2001; Kephart & Chess, 2003; White et al., 2004) and autonomic communications (Smirnov, 2005; Dobson et al., 2006) have focussed on how such systems can automatically adjust their performance and behaviour in response to the changing conditions of their work environments. The goal of this research, involving the whole scale of contemporary computer science from automata theory to artificial intelligence, is to improve and enhance the complex design of modern computing and communication architectures with organization and maintenance capabilities that occur in living systems (Foerster, 1962; Eigen, 1971; Miller, 1978; Eigen & Schuster, 1979; Kauffman, 1993; Camazine et al., 2003) known as self-organization (Heylighen & Gershenson, 2003; Prehoferand & Bettstetter. 2005; Mamai et al., 2006), self-configuration (Hammer et al., 1995; Wildstrom et al., 2005, Gilaberte & Herrero, 2009), self-optimization (Shen & Thomas, 2008; Debrela et al., 2009, Satzger et al., 2009), self-healing (Olsen et al, 2006; Hassan et al., 2006, Samie et al., 2009) and self-protection (Sterritt & Hinchey, 2005; Jarrett, & Seviora, 2006; Dai et al., 2006).

The development of *autonomic* algorithms, protocols, services and architectures for the next generation pervasive Internet that "evolve and adapt to the surrounding environments like living organisms evolve by natural selection" (Miorandi, 2006) is expected.

Formality in contemporary computing means that information processing is both abstract and syntactic and that the operation of a calculus depends only on the form, i.e. on the organisation of the representations which have been considered as deterministic due to oversimplification. Yet, although scientists are equipped today with a whole range of theories and tools for dealing with complexity in nature (Boccara, M. 2004; Sornette, 2004), they miss the real essence of *natural computation*. The more discrete computation pervades into details of natural environments, the more it faces the critical phenomenon of *incompatibility* of state-based computing models discussed in the previous sections to such areas as seismology, meteorology and the stock exchange, (Koestler, 1967). The diverse models and tools for solving differential and recurrence equations, for modelling stochastic processes and power law distributions using cellular automata and networks are operating on Turing machine computer architectures based on the concept of *state*. The latter is central for Newtonian mechanics which describes reality as sets of discrete interactions between stable elements.

However, physics had already a paradigm shift towards quantum mechanics in the past century and natural systems are now seen at their utmost detail to match the wave function equation which describes reality as a mesh of *possibilities* where only an undifferentiated potential describes what *might* be observed a priori to measurement.

Biological systems are even more complex than physical ones and what is needed is a different paradigm such as that of quantum information processing (Prevedel, R. et al., 2007) *embedded* in the underlying computation architecture in order to achieve the vision of truly autonomic computing and communications.

*Networks* are seen as the general organizing principle of living matter. A recent IBM report in Life Sciences (Burbeck & Jordan, 2004) referred to a *Science* article which made this conclusion (Oltvai & Barabasi, 2002). Indeed, this finding originates from Rashevsky's biological topology (1954) and Rosen's relational biology (1958) which became the foundations of modern theoretical systems biology. The network concept in life was then reinforced 20 years later by the autopoietic theory of Maturana and Varela (1974) and by Miller's living systems (1978).

Almost a decade before carrying out the first computer simulation experiments of autopoiesis, (MacMullin & Varela, 1997), Varela and Letelier tested Sheldrake's theory of morphic resonance in silicon chips using a microcomputer simulation, (Varela & Letelier, 1988). Briefly, they let a crystal grow as a microcomputer simulation. It was expected that if the morphic field theory were correct, the synthesis of the same crystal pattern would be accelerated after millions of iterations. Yet, Varela and Letelier found that there was no detectable acceleration of the growing process at all. They concluded that either Sheldrake's hypothesis was falsified or that it does not apply to silicon chips. However, there might be other explanations of this result. One of them could be that conventional Turing machines were used to simulate the above experiments. In this case, new computation models beyond the TM concept are needed to address such issues as autopoiesis. One of the key technical questions to be solved is "to what degree and in which sense life is digital" (Mesarovic & Sreenath, 2006).

Several issues arise in the investigation of non-Turing computation:
(i) What is computation in the broad sense? (Copeland, 1996; Lamport, 2008; Horswill, 2008)
(ii) What frames of relevance are appropriate to alternative conceptions of computation (e.g. natural computation, nanocomputation), and what sorts of models are needed for them? [Some of the most common abuses of mathematics in biology are "situations where mathematical models are constructed with an excruciating abundance of detail in some aspects, whilst other important facets of the problem are misty or a vital parameter is uncertain to within, at best, an order of magnitude." (May, 2004).]
(iii) How can error, uncertainty, imperfection, and reversibility be fundamentally incorporated into computational models?
(iv) How can new physical processes (molecular, biological, optical, quantum) be systematically exploited for computation? (MacLennan, 2006)

The above questions were posed, discussed and partially answered in a different context already. In particular, computing with noise (iii) is a well developed field. Some types of noise reduce the computation power, whereas others tend to increase it. For instance, a model for analog computation in noisy environment was introduced in (Maass & Orponen, 1998). It was shown that it can recognize only regular languages for a specific type of noise. Analog neural networks with Gaussian-like noise appeared to be also limited in their language-recognition power to certain languages (Maass & Sontag, 1999). These results are in sharp contrast with those in a noise-free environment where analog computational models are capable of simulating Turing machines and recognizing non-recursive languages when dealing with real constants (Siegelmann, 1999). In (Ben-Hur et al., 2008), the authors propose a language recognition model based on probabilistic automata (Kleene, 1955; Rabin, 1963) to model analog computation in noisy environments. The main result is a generalized reduction theorem that implies that under very mild conditions the computational power of the automaton is limited to regular languages. On the other hand, it was shown that adding a certain amount of environment noise can improve epistasis models of genetic data in a computational evolution system (Greene, et al., 2009).

In ecological systems a highly specialized organism can fail to thrive as the environment changes. By introducing numerous small changes into training data, i.e. the environment, during evolution, selection can be driven towards more general solutions. This improves the computational power of the evolution system when modest amounts of noise are used. Furthermore, this method of changing the environment in which fitness is evaluated with small perturbations fits within the computational evolution framework and is an effective method of controlling solution size for problems where the data are likely to be noisy. Similar results could be expected (or not) in real life evolving ecologies such as pervasive and cloud computing environments.

However, the above issues have been usually examined in the classical exo-scientific way, where the observer is not part of the system. To obtain the whole picture within an integral approach, however, questions i) - iv) need to be studied also "from within", following an endo-physical (Rössler, 1987/1998; Kampis, 1993/1994), endo-cybernetic (Matsuno, 1998; Thomas, 1998; Dubois, 1998) and endo-technological (Svozil, 2005; Diebner, 2006; Seaman, 2008; Rudolph, 2009) paradigm of virtual worlds going deeply inside and far beyond W3 (Berners-Lee, 1989). Then both views, internal and external, can be combined to deliver a consolidated model of the system. A possible starting point for this investigation allowing a simultaneous formal exo- and endo-description of living systems was proposed in (Nikolaeva-Hubenova, 2001). Thereby, the cognitive aspects of this study should be examined in the light of general systems theory (Bertalanffy, 1968) and a unifying radical evolutionary constructivism (Bateson, 2002; Foerster, 2002; Glasersfeld, 2007). In particular, mathematics and logic should be practically re-invented reflecting genetic epistemology from the very beginning (Piaget et al., 1960; Piaget, 1965-2007; Glasersfeld, 2007) to face a new *bio*-logical reality (Parker et al., 2004) in the playful manner of Richard P. Feynman (Feynman et al., 2000; Feynman, 2005a/b) with a scent of Dewey's naturalistic empiricism (Dewey, 2008). Finally, one should pay regard to the dichotomy between comprehension and interpretation of such an integral model of the living world, because "we explain through purely intellectual processes, but we understand through the cooperation of all the powers of the mind activated by apprehension" (Dilthey 1894, p. 172).

Integral Biomathics is a new approach towards answering these questions and towards shifting the computation paradigm closer to the domains of quantum physics and biology (Baianu, 1980, 1983, 2004) with the ultimate objective of creating artificial life systems that evolve harmoniously with natural ones. This new discipline is particularly interested in four essential *HOW* questions (implying also the Aristotelian *WHY*):
- How life and life-like properties and structures apparently emerge?
- How abstract ontological categories and semantic entailments emerge in living systems?
- How cognitive processes emerge and evolve in natural systems?
- How life related information is transferred through space and time?

A starting point in this quest is the definition of a theoretical framework where the understanding autopoiesis plays a central role. Kawamoto's extensive definition of autopoiesis can be described as follows (Kawamoto, 2000):

> "An autopoietic system is organized as a network of processes for production of elements. Then, (i) the elements of the system become the components only when they re-activate the network that produces these elements, and (ii) when the sequence of the components construct a closed domain it constitutes the system as a distinguishable unity in the domain in which they exist."

The main difference between Kawamoto's definition of autopoiesis and the original one by Maturana and Varela lies in the second part of the definition (ii). It distinguishes between the living system *itself* (German: "sich" or the "self"-part of the process definition) constituted as the network of productions from the *self* (German: "Selbst" or entity of the system as epistemological distinction such as e.g. awareness of a social group) as the closed domain in the space.

According to Kawamoto, this extension makes it possible to represent the aspect that the entity *Selbst* (self, syntax) changes while *sich* (itself, semantics), which ultimately represents its self-awareness, is maintained much like schizophrenia. Kawamoto developed this extension from psychiatric perspective. He argues that this distinction is ambiguous in Maturana's and Varela's original definition and that it causes misunderstanding of autopoiesis. Nomura's formal model of autopoiesis in (Nomura, 2006) is actually affected by the above aspect. The organization is closed and maintained in a specific category and the structure is open and dynamic in a state space.

Indeed, the first part of Kawamoto's definition of autopoiesis is more precise than the original one of Maturana and Varela. It provides an initial condition which can be regarded as the "birth" of the living system. The second part of the definition also identifies a more distinct characteristic than the original one (Maturana & Varela, 1980). Here, the verb "construct" implies that the production of the closed system takes place permanently, i.e. in every single moment, so that the system does not die and then revive again and again. Thus, the distinction from the environment is always present and includes the processes of metabolism and repair which maintain the development of the living entity and its equilibrium/exchange with the environment (homeostasis).

Living systems are open in physical spaces. But autopoiesis requires closedness of organization in living systems. This implies that openness – closedness (enfolding – unfolding or syntax – semantics) lie within observer's physical perspective level and another level beyond it, as described in Nomura's model of two levels illustrated with the relations in figure 3. Indeed, the layering of perspectives has also the dimensions of Maturana, Varela and Luhman who defined the categories of first, second and third order autopoiesis starting from molecules and cells, moving up through organic systems and individual beings to species and social organisations.

It could be asked whether Nomura's model represents an orthogonal view to the classical model of autopoiesis while containing subsets or overlays of the three orders of autopoiesis defined there. The above problem is not explicitly dealt with in Nomura's paper (Nomura, 2006). However, the existence of an isomorphism between operands and operators, the necessary condition of completely closed systems, is implied from the orthogonal view mentioned in (Soto-Andrade & Varela, 1984). This perspective appears also when Rosen's idea is re-considered (Rosen, 1991). Nevertheless, some hardline philosophers including Kawamoto argue that since the view of the relation between inputs and outputs in the system is the one of the external observer, it does not clarify the organisation or the operation of the productions in the system itself, (Nomura, 2007).

Consequently, any description of this level is impossible with the current mathematics at hand. Nomura reckons (pers. corresp. Dec. 2006) that this impossibility implies the difference between the perspectives of quantum physics and autopoiesis regarding the role of the observer, (Toschek & Wunderlich, 2001). The two theories are separated and limited either by their dimensional scale or by the available and applied formal means for description and analysis, or by the present day human perception of reality. At the other end of the physical scale, however, gravitation acts in a similar way as autopoiesis against thermal equilibrium and the homogeneity of systems. The Earth (Lovelock, 1979) and the Universe (Alfvén, H., 1990; Talbot, 1991) appear to be self-organizing just as life and human mind are (Barrow & Tipler, 1986). It is therefore legitimate to ask why there should be any difference in the general principles for organization of matter and life at the micro and the macro scales, a question that initiated the quest of a *unifying field theory* (Boscovich, 1758; Maxwell, 1865; Tesla, 1892a/b; Tonnelat, 1955; Sakharov, 1968; Griffith, 1968; Puthoff, 1989, Haisch et al., 1998; McTaggart, 2003) continuing today within the realms of such frameworks as the M-theory (Duff, 1996; Chalmers, 2006) and the BSM-theory (Sarg, 2006). [Einstein's seminal contributions in this area are indisputable (1905-1915); yet, the author does not share the viewpoint of this line of research entirely, despite a recent report about his reconciliation with the concept of ether (Kostro, 2000).]

Yet, each perspective has its own rules and frame of relevance as MacLennan states, so that it should rather be asked at which level autopoiesis can be defined in the classical way. At the cellular level that could be a good model, but at the molecular level quantum effects such as non-locality or entanglement (Aspect et al., 1982) begin to play an essential role. Since Rosen claimed that a material system is an organism if and only if it is closed to efficient causation, the above evidence leaves open the question *at which level a system can be defined as open or closed* and if a strict separation of concepts can be provided. The whole circle of questions around these definitions is not complete at the current state and their formulation and answer will require further study.

One aspect in modern biology and physics, however, appears to be common: *relativity*. Newtonian physics deals with closed systems and treats space and time as *absolute* variables. The same is true for ordinary quantum theory.

However, proportions, relations, correlations, interrelations and coherences between the different perceived dimensions and parameters of the Universe appear to manifest the most essential property of Nature. This line of thought has been traced all throughout the works of Pythagoras (Guthrie, 1987), Leibniz (1684-1703), Kant and Blumenbach (Lenoir, 1989) to 20$^{th}$ century Peirce (1901), Einstein (1905-1915), Rashevsky (1954 ff.), Rosen (1958a/b ff.), Nottale (1993) and Barbour (1989-1999). It was realized that position and motion have no meaning except relative to other entities in the Universe. Similarly, time has no meaning either, except to some processes of change in this Universe.

A direct conclusion of Einstein's theory of general relativity is that gravity is a result of the relational nature of space and time. That life is the result of other relations, those of functional organizations and the structures which realize them, is also a conclusion of relational biology (Rosen, 1958a, 1959, 1962a, 1991). Hence, it could be asked if there is any connection between gravitation and life (Smolin, 2003). Both domains have similar bases: *fields* and *relations*.

Gravitation is the weakest known physical field interaction, (Misner et al. 1973). Morphic resonance is presumably a subtle field which interacts with all other (physical) fields at all levels of reality (Sheldrake, 1981). Furthermore, if there is logic applicable to cosmology, it must be one that depends on the observer, a *relational logic* (Smolin, 2000). The same argument explains the different perspectives and domains of biology which are often separate from each other and make them appear to belong to different science disciplines.

"Relational variables are created by the system itself, as it evolves. They do not exist a priori, but are defined in a context of relationships created by the dynamics of the system" (Smolin, 2003). Thus, the theories of geometrodynamics (Wheeler, 1962-1964) and quantum gravitation (DeWitt, 1967a/b/c), which introduce the hypothesis of *reprocessing* the final state of the Universe to solve the Big-Bang singularity problem, represent another interrelationship between cosmology, biology and computation. This circuit attempts to explain the (re-)creation and the evolution of the Universe at the Omega/Alpha Point. Accordingly, if the calculation of the physical constants – assumed are *variable constants* (Barrow & Tipler, 1986) – is shifted by small random amounts, this process delivers a *population of evolving universes* analogous to the population of a living system.

This is then called the law of *cosmological natural selection*. However, it is not clear yet what kind of law is going to be natural selection in the context of autopoiesis. Most probably, this is the kind of *law without law* (Wheeler, 1983), a natural pattern or habit (cf. Fig. 3) where quantum entanglement and morphogenetic fields among the multiplicity of universes may be responsible for changes in the past. The manifestation of innate *semantics* and *semiotics* of an enfolding/unfolding holographic Universe is open (Bohm, 1980; Talbot, 1991) and probably related to the *maximum entropy production principle* (Salthe, 2002: Dewar, 2003, 2005).

In this way, *networks*, *fields*, *relations* and *semantics/semiotics* provide the common base in science for understanding the emergence, organization and evolution of life in the Universe. The most interesting result of the research surveys presented in this paper is the parallel in the discussions about the formalizations of *life* (♣), *thought* (♣♣) and *computation* (♣♣♣) which the author regards as the fundamental questions of a new scientific discipline.

Efficient and reliable service provision is the key to next generation complex communication systems (Turner et al., 2004). Yet, current efforts to develop autonomic computational and communication systems that claim to mimic biological ones are paradoxically continuing the system design line of using discrete mathematics for modelling conservative physical systems to produce deterministic self-* automata, (IBM, 2001).

These engineered systems are expected to evolve and adapt according a limited set of production rules to maintain continuous system equilibrium. Such behaviour can be described, for example, by using the equations of classical dynamics. Unfortunately, this is a fairly limited case in Nature. Most physical and chemical processes exhibit alteration, irregularity and contingency along with consistency, orderliness and pattern formation. Besides, biological systems are not conservative, but *dissipative* with fluctuations, unsteadiness and irreversible processes playing a central role, (Nicolis & Prigogine, 1989). In addition, contradictions and paradoxes (as the ones known in quantum mechanics) cannot really be banned from natural science, for they are an integral part of life and cognition (Pietschmann, 1980). The more classical engineering and computation enter the domains of natural science, the more they face these phenomena and anomalies. Science does not know life systems well enough to make the claim of being able to create them error-free. Artificial life and intelligence cannot be compared with their natural opposites; they are based on different principles.

*Integral Biomathics* is envisioned to discover and establish new relationships and deliver new insights into the interaction and interdependence between natural and artificial (human-created) phenomena for a number of scientific fields. It is expected to invent and develop new mathematical formalisms and provide a generalized epistemological framework and ecology for symbiotic research in life, physical, social and engineering sciences.

The author anticipates that Integral Biomathics and its new way of thinking will have an impact well beyond the domains of natural sciences on such disciplines as biology, physics and medicine, computer science and engineering, information technology, electrical and mechanical engineering, bioengineering, material sciences including nanoscience and nanotechnology, aeronautics and astronautics, civil and environmental engineering, architecture and design, energy and earth sciences (geology, ecology, meteorology, geophysics, seismology), business and education (global and political economy, economics, management science and engineering, finance and marketing, organisational behaviour, decision and risk analysis, production strategy and policy, operations management) and the humanities (visual and performing arts, history, languages, literature, psychology and philosophy).

This new kind of science is going to be another challenging mountaineering experience for human intellectual development. Yet, the author remains optimistic, for history of science also knows of other unusual discovery pathways which proved to be successful in the long run (Crick, 1988).

## 8. Summary

This paper continues a line of research set up by previous work in relational mathematical biology (Rashevsky, 1954, 1960a/b, 1967b, 1969a/b, 1970; Rosen, 1958a/b, 1959, 1961, 1962a; Chytil, 1977). Its purpose is to point out (again) that "various relations within organisms … are outside the scope of metric mathematical biology", (Rashevsky, 1958).

If the intention of recent research roadmaps in computational systems biology (Sleep, 2006) and autonomic communications (Carreras et al., 2006) is to enable the free knowledge, technology and methodology transfer between the domains of analytic life sciences and synthetic engineering sciences (computer science, robotics, telecommunications, etc.), both sides should be aware of this fact. Such roadmaps should consider the limits of the current state of technology and formal models when the question is about exploring and using as a model the complexity of the organic world.

There is something beyond the Galileian/Cartesian/Newtonian approach to science in this case. The reason is that that "*ultra-complex* living systems cannot be simulated with any finite number of such robotic, mechanistic, or complicated computer models … by utilizing numerical computation algorithms which are based on *recursive* functions", (Baianu, 2006).

Besides, "… in contrast to causality on one level there is a biological uncertainty principle of causality across levels – akin to Heisenberg's principle in physics – embodied in the concept of a *bounded autonomy of levels* (BAL)" (Mesarovic & Sreenath, 2006). Another dichotomy to keep in mind is development vs. evolution. Development is knowable, scripted, while evolution is the effects of historicity. Development can be theorized, evolution, not (Salthe, 1993).

Roger Penrose pointed out that Turing's own work showed the limitations of computability (Penrose, 1994). He also questioned what Turing would have made of the physics of the brain in 1938, if he had taken seriously the idea expressed in his ordinal logics that the mind does something incomputable in seeing *intuitively* the truth of the Gödel statement. Perhaps the time for revisiting and revising the foundations of computational logic beyond classical issues such as the stability and halting problems in neuron networks (Garzon & Franklin, 1989) has come now again.

Integral Biomathics – the name is not important in the end*,* but the principle – is about non-reductionism and relativism in science, thus unifying the endo- and exo-physical approach to system biology. The first step towards an 'everything' naturalistic science of large systems which is based on cohesion and synergism rather than on dichotomy and discrimination between the constituting disciplines, could be made by computer science.

Three hundred years after its invention, the binary number system conquered the world for the main reason that many electronic components exhibit *two stable states* (Lautz, 1979). The advantages of modern computing are doubtlessly enormous, although living systems are not using this code. Hence, the question is: should non-living matter at the digital switch scale define the long-term matters in science when biology has its turn?

The term 'pre-Newtonian' was used by Rashevsky to qualify the mid 20$^{th}$ century development of mathematical biology which was basically dealing with a collection of isolated phenomena, but not able to explain life in its integrity (Rashevsky, 1954, p. 320). Now, over 50 years later, thanks to computer-aided research scientists can process more data. Yet, they still encounter the lack of general principles in biology and get lost inside the information flood. In addition, the "direct application of the physical principles, used in the mathematical models of biological phenomena, for the purpose of building a theory of life as an aggregate of individual cells is [still] not likely to be fruitful" (*ibid.*, p. 321).

Hence, the term 'post-Newtonian' is used in this survey with a quadripartite emphasis on: i) the missing fundamental set of natural laws in biology, as they are known in classical Newtonian physics; ii) the missing development in biology towards quantum-relativism as this happened in the 20$^{th}$ century physics; iii) the necessary paradigm shift towards a revised modelling relation in science with generalization of biology as shown on Figure 3; and iv) the necessary *belief* changes about the nature of true biological computation and communication beyond the classical Turing Machine concept which reflects the world in terms of Newtonian mechanics.

The author claims that a post-Newtonian view into the logos of bio should answer such questions as:

- *What is computation?* – within the biological context, because there is "no computer into which we could insert the DNA sequences to generate life, other than life itself" (Noble, 2010).
- *How useful is computation?* – for living systems, where "usefulness" is studied from the viewpoint of the entity performing the computation.
- *To what extent can a computation be carried out?* – in an organism or an ecosystem, with the available resources (energy supply, time, number of 'computing' elements, etc.).

To face all these issues, it is mandatory to revise the conceptual framework of contemporary computing and communication theory, rather than addressing other issues such as computability which are essentially irrelevant to biology. Since historical contingency prevents classical Turing computability in dissipative systems, including the biological ones, alternative theoretical approaches to defining *biocomputability* in line with those in (Hogarth, 1994; MacLennan, 2003a) are required. The author regards living systems as morphogenetic fields delivering structural-functional patterns by means of autopoietic networks and thermodynamic organization involving some 'hidden' semantics (Bohm, 1952) within and among the structural elements. In particular, this interrelationship can be described by representing organisms as built of *n*-ary relations (Rashevsky, 1959) or *n*-placed predicates (Rashevsky, 1965b). It is essential, however, to identify and validate optimal design principles, e.g. optimal minimal vulnerability (Martinez, 1964), in such models.

## 9. Conclusions

There are numerous approaches that can be used as a base for the integral mathematical description of living systems such as category theory (Rosen, 1958b), set theory (Rashevsky, 1959, 1961b, 1967a), graph theory (Rashevsky, 1955b, 1956; Rosen, 1963; Rashevsky, 1968), predicate logic (Rashevsky, 1965b) or combinations (Rashevsky, 1958) and cross-products of them (Comorosan & Balanu, 1969; Baianu et al. 2006). A new kind of *bio-logic* might be also invented (Elsasser, 1981). Yet, these formalisms should be capable of addressing such phenomena as variableness, fuzziness, uncertainty and superposition using a new sort of calculus.

A first step towards practical realization of the Integral Biomathics approach could be its application in the formalization of web services (Turner, 2005) or virtualization of IT infrastructures (Rudolph, 2009). The logical frame of this approach is given by the dynamic relational nature of processes in nature where even relations between entities and networks of processes that maintain them are a subject of syntactic replacement and semantic encoding and decoding.

Of course, any other formalism such as lambda calculus could be also used to describe (parts of) the phenomenological framework of living systems. It is essential, however, that the applied mathematical foundation should be capable of truly reflecting the realized relational characteristics of the living system as an open interacting and developing entity, which imposes the tough requirement that the formalism itself must exhibit the same qualities, and not fall back on postulates which do not belong to the system model. Therefore, it is necessary to begin this journey with the study of the mathematical formalisms. The researcher is asked to step back and investigate his/her tools and methods at hand and then to select, adjust and develop those which are appropriate for the studied matter. Automatism in doing science is not advisable. This holds for any other analytic or synthetic discipline attempting to approach living systems in a mechanistic may.

The empirical methods known from the era of Aristotle and Galileo which dominate natural sciences and engineering today are based on the thesis that a system can be studied if it is made distinct and limited, ergo definite, from its environment, if it is disassembled and made quantifiable in (usually discrete) units in order to be calibrated, compared, ergo measured and related, and classified within other systems of the same or similar kind. These methods have been successfully applied for centuries in moulding the world to match human need. There is nothing wrong with creating artefacts and engineering the world. Yet, this should be done in a *response-able* way.

Artificially designed conservative systems experience problems with their environment when placed inside open natural ecosystems. The problem is *adaptation*. A good example is drug design to combat virus diseases. No matter how complex and sophisticated a medicament is, it can never be as good as the virus in the long run, unless it becomes a living system itself that learns *how* to neutralize its enemy. The only system that has been identified as being capable of handling this problem is just another organism. In order to advance in this field, it is necessary to change the traditional approach to medicine. Living systems are capable of learning. The whole is more than the sum of its parts in the living world. Children know that a living system is not living anymore if it is broken and dismantled into bits and pieces. Constructivism is of little or no use in such cases a posteriori. Yet, mathematics knows other methods of studying systems such as comparing surfaces *by analogy* without taking them apart. Certainly, an integral approach to both perceiving and realizing natural systems is required to elevate science to the next stage of development.

Popper's doctrine of falsifiability states that only those theories that are testable and falsifiable by observation and experiment are properly open to scientific evaluation. Integral Biomathics is yet to prove that it satisfies this criterion.

The formalization of autopoiesis and the engineering of living systems represent two non-trivial problems in modern science (Rosen, 1999; Chu & Ho, 2006; Nomura, 2007). It is appropriate to question to which extent traditional logical, metrical and statistical approaches are in place here. On the other hand, hierarchical, relational and holistic approaches to biology such as those of Thompson (1917), Uexküll (1920b), Rashevsky (1954 ff.), Rosen (1958a ff.), Salthe (1985 ff.) as well as Salthe and Matsuno (1995), appear more appropriate in a fuzzy domain. Therefore, a different attitude and perspective towards the study of natural systems both from inside and outside could be quite helpful.

This paper provided a critical survey of present day research in a number of areas of natural sciences and engineering. It is an attempt to discover some forgotten and neglected ideas and theories of the past (Boscovich, 1758; Ehrenfels, 1890; Dilthey, 1894; Lodge, 1905; Thompson, 1917; Rashevsky, 1933-1970; Gurwitsch, 1944; Franck, 1949; Rosen, 1958-1999; Sakharov, 1968; Bateson, 1972; Musès, 1972-1994; Sheldrake, 1981; Elsasser, 1981; Gariaev, 1994; Rössler, 1987; Barbour, 1999; Sarg, 2006 and others) that could benefit the future in the sense of Whitehead (1933), Kuhn (1962-1963) and Sacks (1995), to raise (again) important issues demanding solutions and to outline some promising approaches along the way towards a third millennium discipline of *Integral Biomathics*.

**10. Outlook**

Over 50 years ago Rashevsky pointed out that "the relation between physics and biology may lie on a different plane from the one hitherto considered" (Rashevsky, 1954). He supposed that while the physical phenomena are the manifestations of the metrical properties of the four-dimensional universe, the biological ones may reflect some local topological properties of that universe. Realizing that biology needs stable fundamentals such as those of physics since Newton, his goal was to develop a set of principles which connect the different "physical phenomena expressing the biological unity of the organism and of the organic world as a whole".

Thus, studying the theory of organic form (Thompson, 1917), Rashevsky proposed the *Principle of Maximum Simplicity* as determining the sizes of the different parts of an organism and thus its overall shape (Rashevsky, 1943a, b, c; 1944; 1948). He postulated that "the structure of an organism is determined as the simplest possible physical design which can perform the necessary function with the required intensity" (1954). Rashevsky realized that when observing the phenomena of biological integration, there is evidence not of continuous or discontinuous quantities, but of qualities expressed in complex and often implicit relations that are closer to topology and the theories of groups and categories, rather than to number theory, algebra or other branches of mathematics. He observed that such analogies go much deeper into the realm of living when not merely the structural, but also the functional relations are taken into consideration. Yet, other aspects such as energy exchange increase the relational complexity of living systems. Thus, Lotka has suggested the *Principle of Maximal Energy Flow* as a general physical principle in biological phenomena (Lotka, 1922a; Lotka 1922b) and Rashevsky has attempted to elaborate this principle mathematically, yet without success (Rashevsky, 1948).

Later this research has been further developing the thermodynamic perspective on life while trying to explain evolution (Odum & Pinkerton, 1955; Odum, 1983; Ulanowicz, & Hannon, 1987; Brooks & Wiley, 1988; Salthe, 1993; Schneider & Kay, 1994; Chaisson, 2001; Lorenz, 2002; Lineweaver, 2005; Chaisson, 2005) towards a Unified Theory of Biology. Also, information theory and self-organized criticality in non-equilibrium stationary states (Dewar, 2003), as well as autocatakinetics and spontaneous order (Swenson, 1997-2000) were involved to deliver even a more complex picture of the living systems phenomena. Recently, this field known now as the *Maximum Entropy Production Principle (MEPP)* blossomed in a series of elaborated works (Salthe, 2004, Dewar, 2005, Whitfield, 2005; Sharma & Annila, 2007; Jaakkola et al., 2008; Kaila & Annila, 2008; Mahulikar & Herwig, 2008; Chaisson, 2008).

But for all that, despite the numerous and often justifiable criticism on the biofield theories such as (Stanger, 1999) aiming at debunking them as non-scientific vitalistic approaches to life sciences, there is a significant body of evidence in both theoretical and experimental research in these areas including biological, neural, morphogenetic and human energy fields (Gurwitsch, 1944; Griffith, 1968; Sheldrake, 1980; Dosch, 1999; Korotkov, 2002; McTaggart, 2003), bioenergetics and bioelectronics (Szent-Gyorgyi, 1957, 1968), biophotonics and bioelectrodynamics (Popp, 1984; Popp, 1976; Popp et al., 1989; Mae-Wan et al., 1994; Beloussov & Popp, 1995; Voeikov & Naletov, 1998; Voeikov, 1999; Chang et al., 1998; VanWijk, 2001; Popp, & Beloussov, 2003; Beloussov et al., 2006), wave genetics (Gariaev, 1994, 1997; Gariaev et al., 1999), bioinformation and biocomputation (Gariaev et al., 1991a/b; Berezin et al., 1996; Gariaev et al., 2000) that cannot be simply ignored today.

A major paradigm shift in biology, medicine, natural sciences, computation and engineering is already on the way. "For a conceptual breakthrough, a new paradigm is needed for a complex biological phenomenon beyond the current parameters of networks and of systems", (Mesarovic & Sreenath, 2006). A new fundamental understanding about the role of biology in analytical natural sciences and synthetic engineering sciences has to be set out. Where could be the join between biology and computation? Dissipation was found to be an integral part of computation (Porod et al., 1984). Despite the "lack of dissipation" in system-biological models of the past (Boogert et al., 2007), recent results in molecular biology and computation suggest even more complex relationships between information content and energy dissipation in organic structures. Evolution itself may appear to be the result of entropy in the universe (Brooks & Wiley, 1988). Thus, in addition to information, a DNA molecule also stores energy using adenosintriphosphate (ATP) as fuel, thus providing both the input data and all of the necessary energy for a molecular automaton (Benenson et al., 2003). This allows the processing of an input molecule of an arbitrary length *without* external energy supply, which is of course not the case in classical Turing Machine computation models. Another evidence for a quite different base and level of biocomputation is the remote holographic type replication of genetic information in the *DNA-wave biocomputer* (Gariaev et al., 2000).

Therefore, the helical structures in the evolving modelling relations for science depicted in figures 4 and 6 imply the creation of a profoundly new theory for biology and physics based on the profound understanding of the complex relationship *space-time-matter-life-cognition-consciousness* involving the processing, communication and transformation of energy and information.

The form of a helix may have a deeper meaning along the line *DNA/RNA structure – wave genetics – string theory – BSM theory* (Watson & Crick, 1953; Gariaev, 1994; Greene, 1999; Sarg, 2006) that can help understanding the basic building blocks of Nature and finding the ways of their discovery.

Taking this into account, the following research areas are of major interest for Integral Biomathics:

**1. Fields**: The most risky and challenging part of research is concerned with the exploration of the *physical base* of biocomputation in terms of regular assemblies of structural elements and functional patterns (Turing, 1952; Gierer & Meinhardt, 1972; Rosen, 1991; Walace, 2000; Wolfram, 2002; Andras, 2005) excited by the oscillatory character of the underlying processes (Gurwitsch, 1910-1944; Thompson, 1917; Lotka, 1920; Weiss 1939; Frank, 1949; Sheldrake, 1981; Greene, 1999) allowing the evolution of stable nonlinear systems through moving equilibria (Spencer, 1897; Lotka, 1921; Alfvén, 1942, 1990; Zecevic & Siljak, 2003).

For this research new findings about the fundamental nature of the physical continuum in support of unified field theories (Boscovich,. 1758; Tesla, 1892b; Sakharov, 1968; Puthoff 1989, 1993, 2002; Sarg, 2006) addressing the explanation of transformations between energy and information underlying the emergence and evolution of life and consciousness (Penrose, 1989; Hameroff, 1998; McFadden, J. 2001; Sarg, 2003) will be essential.

The expected results of this research are findings in support of a new *biological information theory* which complements the classical one into an *integral information theory* for both artificial and natural systems.

**2. Relations:** Natural computation is relative. Relational variables are created by the system itself, as it evolves (Smolin, 2003). Organisms can be represented as *n*-placed predicates or *n*-ary relations (Rashevsky, 1965-1968). There are numerous other mathematical and physical approaches that can be used as an *abstract base* for the symbolic description of living systems such as category theory, set theory, graph theory, scale relativity or combinations and powers of them.

The expected results are the definition a formal *bio-logic* analogous to Elsasser's (1981) and Rosen's (1991) concepts and the development of a relational calculus that depends on the observer (Smolin, 2000). However, this formalism should also be capable of addressing such phenomena as instant response to unpredicted stimuli, variableness, fuzziness, uncertainty and superposition.

**3. Networks:** According to autopoietic theory, a living system as a clearly distinguishable *network* of processes for production of elements constituting and re-activating the network that produces these elements. Thus, biocomputation – being organized as a network (i.e. implying communication) – might be a by-product of ongoing structural coupling (a posteriori) between collections of autopoietic elements, but it cannot be defined as a purposeful task for the solution of a specific problem or class of problems in the way expected from artificial computational systems today. Its fundamental *intent* is not decision making, but adaptation, life maintenance, survival and replication. Any formal description of such a living network is impossible with current mathematics (Nomura, 2007).

The research in this field is expected to deliver new insights in autopoiesis, as well as a new consistent definition of autopoiesis, a new formalism, or both.

**4. Evolving Hierarchies:** Computation occurring in nature always involves implicit semantics and semiotics. Hence, it cannot be formalised in the conventional way using purely explicit (syntactic) and static Hilbert logic. An expected result of this research is the development of a meta-model describing the emergence of dynamic attributed ontologies in terms of *multi-layered patterns* (living system codes), capable of expressing such phenomena as both natural and artificial neuronal activity.

The patterns themselves and their spatio-temporal formation, use and recognition as signals followed by transformation into signs (semiosis) should be explained with higher layers of order studied in the domains of semiotics (Peirce, 1869b, 1903; Uexküll 1940, 1982), physiosemiotics (Deely, 2001) and biosemiotics (Salthe, 1985-2000; Barbieri, 2007).

## 11. Epilogue

The issues presented in this paper are not only a matter of science, but also of natural philosophy. The latter should be involved in shaping Integral Biomathics from the very beginning to guide scientific discovery, and not to review it a posteriori. A few points in this respect deserve special attention here.

Symmetry is a central principle in science, but asymmetry drives the emergence of self-organization and life (Noether's theorem: Noether, 1918). In Hegel's eyes both symmetry and asymmetry (a dyadic system) would be complementary principles giving raise to something new. Dyads are unstable, triads are stable. This legality relates to Salthe's evolving hierarchies concept (1985). The compositional (scale) hierarchy, with its three levels, models stable (symmetrical) situations, but the subsumptive (specification) hierarchy models emergence, – {physical dynamics → {chemical connectivities → {biological forms → {sociocultural organizations}}}}, – because it is asymmetrical, as it has no explicit upper level and is not prevented from moving to a new level by the anchoring presence of a third level.

Consequently, the interfusion of the natural and formal world on figure 6 will be the result of such a development, creating a new level of organisation when overcoming the sociocultural boundaries. In other words, as the formal and natural worlds interact increasingly more intensely, an intermediate world may emerge between them. This new world – let us call it "meso world" – would presumably be some kind of Noospheric construct (Vernadsky, 1945), less mechanistic than the formal world, but also less vague and generative than the natural world. According to Salthe, new levels in the compositional hierarchy emerge between previous levels, and at the apex of subsumptive hierarchies (Salthe, 1993). Hence, the triad of worlds in this context appears to be not hierarchical, but Peircean – First, Second, Third – that is, a thing in itself, interaction, and habit (Peirce, 1901). However, since the meso world emerges and develops within the limited space-time continuum of human beings, it can be also regarded as a hierarchical construct in cosmological scale, thus being able to interact with other worlds beyond the human sphere of activity. One example would be an ecosystem on a space station which develops autonomously during its exploration mission.

From the time of Descartes, Goethe and Shelley to the present day the idea of the mechanistic nature of life (Loeb, 1912) has received strong support in scientific circles. After all, among all kinds of dissipative structures, the biological ones have devised the most mechanistic components (the bacterial flagella, the "genetic code", etc.). Nevertheless, the mechanical discourse is insufficient to understand any dissipative structures, all of which have capabilities of resilience and of attaining unforeseen states that cannot be assimilated to machine models. Living systems are more complicated because they can recover based on acquired information, and because development leads inevitably to complexity.

On the other hand, computation (as we know it) has an algorithmic, thus mechanistic nature, whereas living forms reveal also thermodynamic and quantum-mechanic properties at different system levels.

Therefore, we cannot simply reduce biocomputation, e.g. by means of DNA molecules, to a set of biochemical reactions that can be easily predicted and programmed in the way Turing Machines were designed. DNA is not the code, but rather (a part of) the database of the living (Noble, 2006-2010). There is also a continuous interaction between DNA and its environment (proteins, cells, organs and the external world) which can be described as "multifactorial and multilevel inward-outward causation". Indeed, biomatter "in a snapshot" behaves like a cellular automaton (Neumann, 1966). Strictly said, it is not the matter alone, but also the processes and functions maintaining it that participate this kind of computation. Thus, protein interaction networks share many features with neural networks (Bray, 2009): i) they are reactive, i.e. deterministic, but unpredictable; ii) there is no limit in the number of predecessors or successors of a node in the network; ii) the signal flow is multidimensional; iii) they are inherently parallel allowing any number of nodes to be processed simultaneously; iv) input and output signals are provided at run-time; v) they behave like digital switching circuit logic while keeping track at variable analog threshold values; vi) they are resilient and learn with time.

However, when development and evolution enter the process the model becomes much more complex, because "there is absolutely no way in which biological systems could be immune from the stochasticity that is inherent in Brownian motion itself" (Noble, 2010). This is probably the one of the reasons why algorithmic efforts to simulate autopoiesis were unsuccessful so far (McMullin & Varella, 1997). Finite state machine models and computing programs have limited applicability in system biology.

Certainly, because of its massive parallelisation, DNA computing is powerful in solving some NP-complete problems. Yet, it does not reflect the real purpose of life structure development, the Aristotelian final cause (telos) or purpose that needs to be found. In physics, the Second Law of thermodynamics calls for entropy production no matter what else happens (at least in the macroscopic realm). Thus, true biocomputation is expected to embrace such a fundamental legality as Carter's and Dicke's Anthropic Principle (Barrow & Tipler, 1986), since it implies some kind of final cause, essential for scientific research.

Nevertheless, a DNA computer (today) is something like a wheel used for motion in translation and not for rotation or for translation through rotation. It does not deliver the 'natural' kind of processing the author is seeking. The danger is that at some point, – perhaps because of its successful operation, – it can be adopted as the *only possible* kind of operation of life structures, thus becoming the same kind of pragmatic dogma in biology which is the Turing Machine in computer science today.

In calculus, there is an expected result to be delivered. The heat death of the Universe is the expected result in physics, but we do not know the expected result of life's development yet. Therefore, it is a good question to ask what exactly is the liveliness characterizing dissipative structures. It is a question of anticipation and autonomy, as well as of flexible reactivity, following the hierarchical relation formula {anticipation → {autonomy → {liveliness}}} (Salthe, 2008, pers. corresp.).

Recapitulating, the ultimate message in figures 1-6 is that *development* will be the key to understanding life, Nature and computation. Being virtually a law of matter, this process characterizes all dissipative systems (immature → mature (in some cases) → senescent). Development could also form the basis of a kind of computation (Salthe, 2008, pers. corresp.). The major trope would be then e.g. {general algorithm → {refinement → {refinement of the refinement → {and so on}}}} with the form being a tree with its root in the most general class. This form and process analysis could be set at *any* system level (Noble, 2006). Thus, Loeb's discovery of the artificial parthenogenesis (Loeb, 1899-1901) can be regarded as a possible base for developmental computation. However, it should be recognised that *development cannot be algorithmic*, i.e. deterministic and predictable – an implication of Gödel's incompleteness theorem (Gödel, 1931), – because "one cannot specify in advance what may emerge in what would have to be a developmental computation of increasing specification {vague → {second order fuzzy → {fuzzy → {crisp}}}}", (Salthe, 2008, pers. corresp.).

This implies a different direction for the development of computation and systems biology from the one commonly adopted today towards *non-algorithmic* computation, such as neurocomputing (Hecht-Nielsen, 1990), neuromorphic engineering (Smith & Hamilton, 1998) and quantum computing (Milburn, 1998). Even if it is true that "causality between events, the temporal ordering of interactions and the spatial distribution of components are becoming essential to addressing biological questions at the system level" (Priami, 2009), multilevel interactions are predisposed to such phenomena as uncertainty, superposition and entanglement. Hence, it cannot be expected that design principles of large software systems today can go beyond improvement of the engineering toolset for such applications as imaging, database search, analysis and simulation in systems biology.

In the past, physics was too much engaged with applying general statistical mechanics for understanding a distinct system state. This approach was inherited in biology and computer science in the past century (Bohm, 1980). However, the Newtonian models currently dominating these disciplines are crisp and fully explicit, while the world is in various degrees vague. But physics continued its development into thermodynamics and quantum mechanics, thus providing new model bases for biology and computation. While thermodynamics connects to development, QM connects to vagueness and internalism. Realising that the one constraint on science, as on all thought, is (conventional) logic, if we wish to go beyond it, the only possibility at present is to address *vagueness,* not algorithms. Finally, a major unsolved problem in biology that will have an impact on biocomputation is the origin of the genetic system. Even Crick finally decided that the DNA must have come from 'elsewhere' in the cell (Crick, 1994), possibly by means of an electromagnetic wave carrier (Gariaev, 1994, 1997)! There is still ongoing experimentation in the RNA world, but biologists hardly like to think about this topic at all and prefer to shift their research focus on developing evolutionary theories instead. This attitude should change in future when pursuing answers on a solid fundament. Biology has the challenging role to answer some of the most error-prone questions of physics today. This is another reason why physics is put within biology on figures 3 and 4. Biologists should finally take the responsibility for their own domain, develop their own theories and research methods and not follow old models that physicists have left behind decades ago or adopt wrong concepts, constructs or metaphors from engineering sciences which suffer the same illness when dealing with complexity.

There is no such thing as a permanent hierarchy of scientific disciplines, except for mathematics which is their queen as Gauss once said. Computational biology of the $21^{st}$ century is set to revise many of the fundamental principles of biology, including evolutionary theory and the relations between genotypes and phenotypes. The study of Integral Biomathics presented in this paper can become a valuable tool to help getting behind the wheel of the required "mathematical insight perhaps of a nature we have not yet identified" (Noble, 2010). Predictable (and programmable), stable discrete (automata) systems are *dead* systems. Autonomous energy supply, reproduction, adaptation to and communication with the environment is not possible. This is how engineering and computation are operating today. The most that can be achieved in terms of the autopoietic definition is virtual or *fake* life. Any solution beyond this is expected to account for development "from within" which attests one of the basic laws of cybernetics and biology that the "relationship between information (J) and thermodynamic entropy (S) is constant (S + J = const.)", (Wiener, 1954).

The above mentioned four research areas, – fields, relations, networks and evolving hierarchies, – provide the framework for a unique realisation of a new generation evolvable biosynthetic computing systems that will possibly deliver some answers along the thread leading towards solving the mysteries of life. This is a challenging goal which is certainly worth working for, because

"… whatever life is or not, it is certainly this: it is a guiding and controlling entity which reacts upon our world according to laws so partially known that we have to say they are practically unknown ..."

<div align="right">Sir Oliver Lodge, *Life and Matter*, 1905, p. 117.</div>

―――――――――

**Acknowledgements:** The author deeply appreciates the valuable help of Prof. Tatsuya Nomura from Ryukoku University (Japan) for discussing questions on the formalization of autopoiesis using category theoretical models, as well as the suggestions and comments of Prof. Leslie S. Smith from the University of Stirling (Scotland) concerning semantics, computation and classical information theory and his final editorship of this paper. He is also very obliged to Prof. emer. Stanley Salthe from Binghampton University (USA) and University of Copenhagen (Denmark) for his in depth review of the paper and stimulating discussions about internalism, semiotics, Peircean philosophy and evolving hierarchical systems. The author wishes to thank specially to Prof. Kenneth J. Turner from the University of Stirling (Scotland) for asking the right questions, to Dr. Dessislava Duridanova from BAS Institute of Biophysics (Bulgaria) and Dr. Vladislav Kolev from the Institute of Physics, London (England) for their critical remarks on system models, as well as to his dear friend and mentor Ursula Saar for her precious time and the exciting discussions about Hegel's and Dilthey's approach to science and the habits of resonance in Nature.

*This paper is dedicated to the memories of my grandfather Joscoe (Joseph)*
*and Royal Navy Commodore A. John M. Donaldson*
*who would be happy to see this work accomplished.*

*PLS*
*Berlin, January 11$^{th}$ 2010*